%% file: main.tex
\documentclass[10pt,twocolumn,letterpaper]{article}

\usepackage[pagenumbers]{cvpr} 
\usepackage{multirow}
\usepackage{tabularx}
\usepackage{array}
\usepackage{cuted}
\usepackage{makecell}
\newcolumntype{C}{>{\centering\arraybackslash}X}
\input{preamble}
\definecolor{cvprblue}{rgb}{0.21,0.49,0.74}
\usepackage[pagebackref,breaklinks,colorlinks,allcolors=cvprblue]{hyperref}

\def\confName{CVPR}
\def\confYear{2027}

\title{Beyond Optical Models: Wave-Aware Gaussian Splatting with Two-Bounce Transport for RF Relighting}

\author{Hailiang Lu$^{1,3}$ \quad Hongyu Mu$^{1}$ \quad Wei Xu$^{1,3}$\thanks{Corresponding author: Wei Xu}\quad Jie Liu$^{2}$ \quad Yao Yao$^{2}$  \quad Xun Cao$^{2}$ \quad Jianhua Zhang$^{4}$\\
$^{1}$ Southeast University\quad$^{2}$ Nanjing University\quad$^{3}$ Purple Mountain Laboratories\\
$^{4}$ Beijing University of Posts and Telecommunications\\
{\tt\small \{hllu, hymu, wxu\}@seu.edu.cn\quad \{liujie,yaoyao,caoxun\}@nju.edu.cn\quad jhzhang@bupt.edu.cn}
}

\begin{document}
\maketitle
\input{sec/0_abstract}    
\input{sec/1_intro}
\input{sec/2_relatedwork}
\input{sec/3_method}
\input{sec/4_experiment}

\input{sec/5_conclusion}

{
    \small
    \bibliographystyle{ieeenat_fullname}
    \bibliography{main}
}

\input{sec/X_suppl}

\end{document}

%% file: sec/0_abstract.tex
\begin{abstract}
Radio-frequency (RF) relighting aims to predict received angular power spectra for unseen transmitter positions. Existing methods often adapt optical rendering frameworks, whose models approximate surface scattering as local specular reflections, thereby neglecting wave effects at RF frequencies. Moreover, fixed base-station deployments often restrict training data to a single transmitter position under a point-source assumption, limiting generalization to unseen incident directions. To address these challenges, we introduce wireless relightable Gaussian splatting (WRGS). To model RF scattering, WRGS first derives a differentiable electromagnetic bidirectional reflectance distribution function (BRDF) from Maxwell equations using the small perturbation method (SPM), accounting for wavelength, material properties, and surface microstructure. To mitigate limited incident-direction coverage, WRGS then incorporates this BRDF into a 2DGS-based two-bounce framework, where inter-surface reflections enrich incident–outgoing direction pairs. Experiments on simulated scenes demonstrate state-of-the-art performance in novel receiver synthesis and zero-shot RF relighting, with up to a 52.6\% reduction in zero-shot mean absolute error compared with the best-performing baseline.
\end{abstract}

%% file: sec/1_intro.tex
\section{Introduction}
Wireless communication plays an important role in society, supporting a wide range of applications, from mobile connectivity and the Internet of Things to autonomous systems and intelligent sensing~\cite{10024766,10054381,You2020,zhao2026electromagneticworldmodel6g}. As wireless networks become increasingly dense and operate in more complex environments, accurately characterizing radio propagation has become a fundamental challenge~\cite{11381448,11104274,10742564}. Recently, the success of 3D Gaussian Splatting (3DGS)~\cite{10.1145/3592433} in computer vision (CV) has inspired its adoption for  characterizing radio propagation~\cite{11355734,11044513}. Despite their promising results, these methods focus on RF field reconstruction under observed transmitter (Tx) configurations and do not readily support varying Tx positions. To the best of our knowledge, generalizing 3DGS-based RF field reconstruction to unseen Tx positions remains an open problem.

\begin{figure}[t]
    \centering
    \includegraphics[width=\columnwidth]{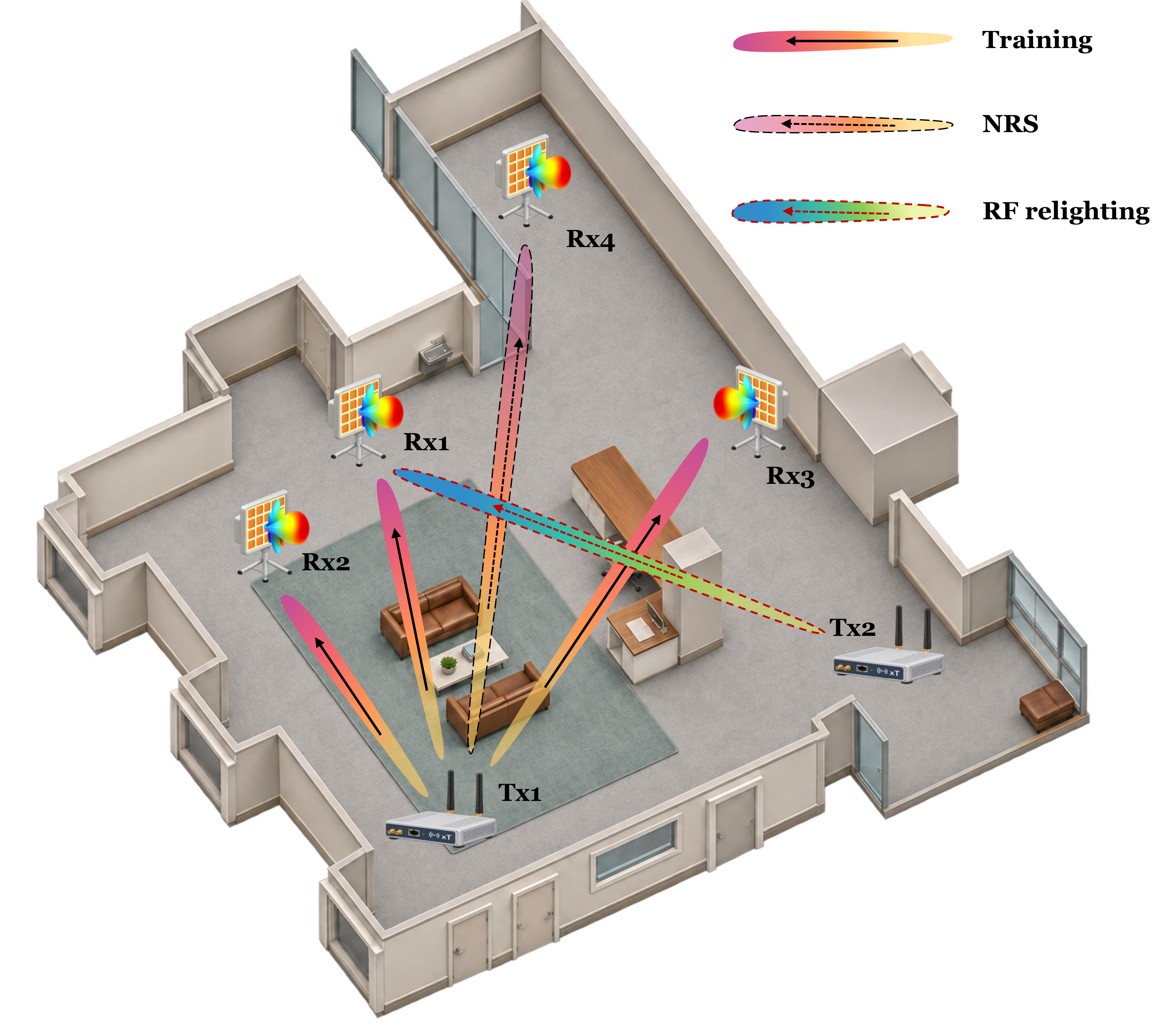}
    \caption{Task definitions of novel receiver synthesis (NRS) and RF relighting. NRS optimizes 2D Gaussians using measurements (Rx1-Rx3) under a fixed transmitter position (Tx1) to synthesize the received angular power spectrum (APS) at a novel receiver (Rx) location (Rx4). RF relighting directly uses the optimized 2D Gaussians to render the received APSs (Rx1-Rx4) with an unseen Tx position (Tx2) in the same scene.}
    \label{fig:task}
\end{figure}

This open problem closely resembles relighting~\cite{990523} in CV. Both visible light and RF signals are electromagnetic waves governed by the Maxwell equations, and changing the Tx position is analogous to changing the illumination conditions of an optical scene. We therefore refer to this task as RF relighting, as illustrated in Fig.~\ref{fig:task}. Motivated by this connection, existing RF relighting methods, including BiWGS~\cite{11720435}, RFIR~\cite{wang2026rfir}, and URF-GS~\cite{wen2026urfgs}, adopt the modeling formulations and rendering pipelines of 3DGS-based optical relighting methods~\cite{Liang_2024_CVPR,10.1007/978-3-031-72995-9_5,Jin_2023_CVPR,ICLR2025_bc97207e,Xu_2026_CVPR,Zhang_2026_CVPR,NEURIPS2025_81144985}. However, the approximations commonly used in optical rendering do not necessarily hold in the RF regime. This limitation is particularly important for RF relighting, which requires predicting how the propagation response of a scene changes when the Tx is placed at unseen positions.

A key limitation of optical surface response models for RF scattering stems from the substantial difference in wavelength scales. Classical Cook-Torrance microfacet models~\cite{Torrance:67,10.1145/357290.357293} are based on geometrical optics (GO) and do not explicitly account for wave scattering from rough surfaces. Since RF waves have much longer wavelengths than visible light, surface variations in many indoor materials can be smaller than the RF wavelength, making GO-based models less suitable for characterizing RF scattering. 

A second mismatch lies in the source excitation conditions of optical and RF relighting. Optical 3DGS relighting methods usually adopt image-based lighting~\cite{Liang_2024_CVPR,10.1007/978-3-031-72995-9_5,Jin_2023_CVPR,11092682}, which provides illumination from multiple directions, whereas wireless transmitters are typically point sources. Under a fixed Tx configuration, each surface point receives direct illumination from a single direction determined by the transmitter-surface geometry, resulting in limited angular coverage and hindering the generalization of learned surface responses to unseen Tx positions. Some optical point-light relighting methods recognize this limitation, but address it by collecting or synthesizing training data under varying light-source positions~\cite{Fan_2025_CVPR,NEURIPS2024_dc72529d,Xu_2026_CVPR,NEURIPS2025_fec4c7df,zheng2026ssdgs,10.1145/3588432.3591482}. However, such pipelines are hardly adopted in wireless communications, where BSs and access points are deployed at fixed locations according to service requirements and cannot be freely repositioned for data collection.

This raises a fundamental question: can physically meaningful RF relighting be achieved merely by adapting representations originally designed for optical rendering? We argue that it requires rethinking how Gaussian representations model RF-specific surface interactions and propagation. Under fixed-Tx excitation, an accurate RF scattering model is insufficient without adequate angular constraints; likewise, richer propagation paths cannot compensate for an inaccurate account of RF surface interactions. Effective RF relighting therefore requires wave-aware scattering and a mechanism that supplies additional incident directions beyond direct illumination.

To address these challenges, we present wireless relightable Gaussian splatting (WRGS), a novel RF relighting framework based on the underlying physics of electromagnetic scattering and propagation, rather than an adaptation of optical relighting formulations. Specifically, we derive a differentiable electromagnetic bidirectional reflectance distribution function (BRDF) using the small perturbation method (SPM) to explicitly account for wave effects in electromagnetic scattering. Furthermore, we introduce a two-bounce framework based on a 2D Gaussian representation to model inter-surface reflections under point-source excitation. By introducing additional incident directions through secondary reflections, this framework increases the diversity of incident and outgoing direction pairs using measurements collected at a single fixed Tx location, thereby reducing the need for measurements from multiple Tx locations. Our contributions can be summarized as follows:
\begin{itemize}
\item We derive an SPM-based electromagnetic BRDF for RF relighting, capturing wave scattering from rough surfaces beyond conventional optical reflection models.
\item We propose a 2DGS-based two-bounce framework for inter-surface reflections, expanding incident-direction coverage under fixed-Tx excitation and reducing reliance on multi-Tx measurements.
\item We conduct extensive experiments on the NIST and WiSegRT datasets, demonstrating state-of-the-art (SOTA) performance in NRS and zero-shot and few-shot RF relighting.
\end{itemize}

%% file: sec/2_relatedwork.tex
\section{Related Work}
\label{sec:related_work}

\noindent\textbf{3DGS in Wireless Fields.}
The explicit primitives and efficient rasterization offered by 3DGS have motivated Gaussian representations for wireless propagation. RF-3DGS and WRF-GS~\cite{11355734,11044513,11258087} represent wireless propagation with Gaussian primitives. GSRF~\cite{NEURIPS2025_5b5618e7} further models complex-valued radiance and transmittance using directional Fourier-Legendre bases. Geometry-aware methods, including GAI-GS, RadCloudSplat, and PC-TGS~\cite{Shen_2026_CVPR,wang2026radcloudsplat,xue2026point}, leverage scene geometry to construct Gaussian representations for modeling multipath propagation and spatial signal distributions. Recent extensions address cross-band propagation, temporal evolution, and reconstruction from imperfect inputs~\cite{10.1145/3795866.3844449,zhang2026terfs,10.1145/3832013}. However, existing methods focus on reconstructing RF fields at observed transmitter positions, leaving the generalization of learned scene representations to unseen transmitter positions largely unexplored.

\begin{figure*}[t]
    \centering
    \includegraphics[width=\textwidth]{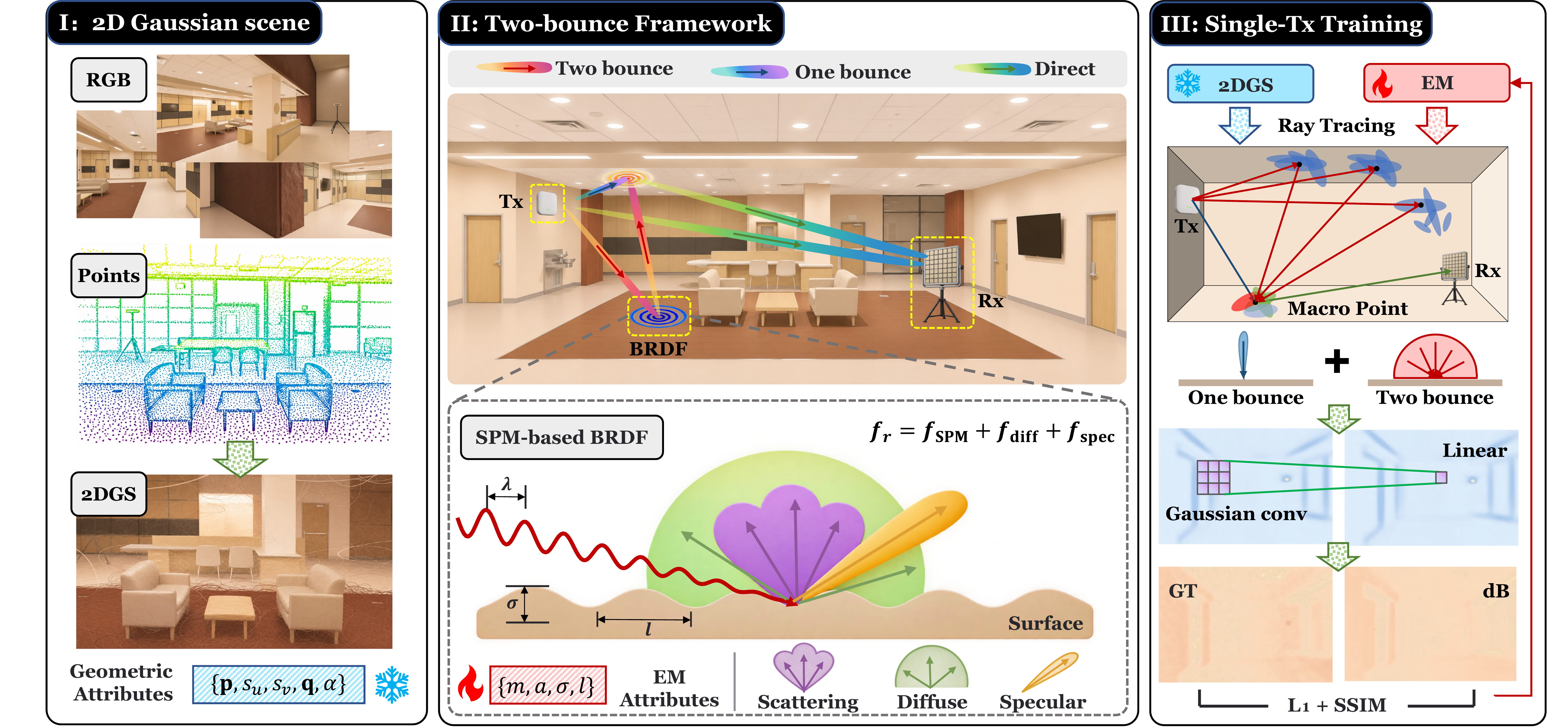}
    \caption{Overview of the proposed WRGS framework. The scene is represented by 2D Gaussians with geometric and electromagnetic attributes. RF propagation includes direct, one-bounce, and two-bounce paths, with surface interactions characterized by an SPM-based BRDF. The model is trained under a single Tx position using ray tracing and Gaussian convolution to predict received angular power spectra (APS), enabling RF relighting under unseen Tx positions.}
    \label{fig:architecture}
\end{figure*}

\noindent\textbf{RF Relighting in 3DGS.}
Recent RF relighting methods have begun to explore generalization to unseen transmitter positions by adapting rendering frameworks developed for optical relighting. BiWGS~\cite{11720435} and RFIR~\cite{wang2026rfir} follow a pipeline similar to that of R3DG~\cite{10.1007/978-3-031-72995-9_5}. Despite differences in projection and scattering parameterizations, both model direct illumination followed by one-bounce scattering. BiWGS employs bidirectional spherical harmonics (BSH)~\cite{11125006} to fit angular scattering responses, making its representation dependent on the diversity of training data, whereas RFIR uses the reciprocal effective roughness (RER) model~\cite{10137406} whose empirical parameterization lacks generality across diverse wireless environments. URF-GS~\cite{wen2026urfgs} adopts the deferred shading and path tracing pipeline of GI-GS~\cite{ICLR2025_bc97207e} to model multi-bounce propagation, but retains a Cook-Torrance~\cite{Torrance:67} BRDF that does not explicitly capture wave scattering from rough surfaces.

%% file: sec/3_method.tex
\section{Method}
In this section, we propose WRGS, a wireless relightable Gaussian splatting framework. As illustrated in Fig.~\ref{fig:architecture}, WRGS integrates an SPM-based BRDF and a point-source two-bounce framework to enable APS synthesis under varying transmitter positions.

\subsection{Limitations of the Cook-Torrance Model}
The Cook-Torrance model~\cite{Torrance:67, 10.1145/357290.357293} is widely used in optical rendering to describe surface reflectance. The BRDF of the Cook-Torrance model is given by
\begin{align}
f_r(\boldsymbol{\omega}_i,\boldsymbol{\omega}_o, \mathbf{x})
&=\frac{DFG}{4(\mathbf n\cdot\boldsymbol{\omega}_i)(\mathbf n\cdot\boldsymbol{\omega}_o)},
\end{align}
where $\mathbf{x}$ denotes the surface point and $\mathbf{n}$ is the unit surface normal. The vectors $\boldsymbol{\omega}_i$ and $\boldsymbol{\omega}_o$ denote the incident and outgoing directions, respectively. The functions $D$, $F$, and $G$ denote the microfacet normal distribution, Fresnel reflection, and geometric masking-shadowing terms, respectively. The model is built upon the GO approximation, which represents rough-surface scattering as a statistical aggregation of local specular reflections.

To illustrate this connection, consider an electromagnetic plane wave incident on a rough surface $\mathcal S$. Under the local tangent plane approximation, the far-field scattering amplitude takes the form
\begin{equation}
E_s(k)=\int_{\mathcal S}\alpha(\mathbf x)e^{ik\Phi(\mathbf x)}d\mathcal S,
\end{equation}
where $k=2\pi/\lambda$ is the wavenumber and $\alpha(\mathbf x)$ is a complex amplitude factor, and $\Phi(\mathbf x)$ is the phase function. The GO approximation assumes that the wavelength is much smaller than the characteristic surface scale $L$, such that the high-frequency asymptotic condition $kL\gg1$ holds~\cite{https://doi.org/10.1029/2000RS002551}. Under this condition, rapid phase oscillations cause contributions from non-stationary regions to largely cancel, whereas contributions near stationary points accumulate coherently. The stationary phase condition, $\nabla_{\mathcal S}\Phi(\mathbf x)=0$, indicates that the difference between the outgoing and incident directions is parallel to the surface normal, corresponding to the specular reflection direction. This result provides a theoretical basis for the local specular reflection assumption underlying the Cook-Torrance model.

However, this argument relies on the scale-separation condition $kL\gg1$. In typical environments, radio waves in commercial wireless frequency bands often do not satisfy this condition because their wavelengths are not sufficiently small relative to the characteristic surface scales. Consequently, phase cancellation in the conventional Cook-Torrance model becomes incomplete in the RF domain, and finite-wavelength interference and scattering effects appear significant. Therefore, microfacet models based solely on high-frequency GO may not fully characterize rough-surface electromagnetic responses in this regime.

\subsection{SPM-Based Electromagnetic BRDF}
In contrast to the Cook-Torrance model, SPM~\cite{https://doi.org/10.1002/cpa.3160040206,FUNG1968125} derives the scattering response from the Maxwell equations and accounts for the wave nature of scattering. SPM treats a rough surface as a perturbation of a planar interface and expands the electromagnetic field as $\mathbf{E}=\mathbf{E}^{(0)}+\mathbf{E}^{(1)}+\cdots$, where $\mathbf{E}^{(0)}$ denotes the planar-interface solution determined by the Fresnel coefficients. Under the small-perturbation assumption, retaining $\mathbf{E}^{(1)}$ captures the roughness-induced scattering while neglecting higher-order terms that incur greater computational complexity.

Consider a plane wave incident on a rough dielectric surface, with polar and azimuthal incidence angles $(\theta_i,\phi_i)$. Let the mean surface lie at $z=0$ and represent the rough surface as $z=h(\mathbf{x})$, where $\mathbf{x}=(x,y)$. The first-order perturbation to the relative permittivity profile is then given by
\begin{equation}
\epsilon^{(1)}(\mathbf{x},z)=(\epsilon_r-1)h(\mathbf{x})\delta(z),
\end{equation}
where $\epsilon_r$ denotes the relative permittivity, and $\delta(z)$ is the Dirac delta function. Substituting this perturbation into Maxwell's equations yields the first-order scattered field:
\begin{equation}
E^{(1)}(r)=\frac{k^2\cos\theta_i\cos\theta_s}{\pi r}QE_0 H(\mathbf q_s-\mathbf q_i)e^{ikr},
\end{equation}
where $(\theta_s,\phi_s)$ are the polar and azimuthal scattering angles, $r$ is the distance from the scattering point to the observation point, $E_0$ is the incident electric-field amplitude, $H$ is the Fourier transform of the surface height, and $\mathbf q_i$ and $\mathbf q_s$ are the projections of the incident and scattered wavevectors onto the mean surface, respectively. The factor $Q$ accounts for polarization and dielectric effects.

The BRDF is defined as the ratio of outgoing radiance to incident irradiance. The incident irradiance and the outgoing radiance of the first-order scattered field are given by
\begin{align}
E_i&=\frac{P_i}{A}=\frac{|E_0|^2A\cos\theta_i}{2Z_0A},\\
L_o&=\frac{dP_{o}}{A\cos\theta_s d\Omega_s}=\frac{r^2\langle|E^{(1)}(r)|^2\rangle}{2Z_0A\cos\theta_s},
\end{align}
where $A$ is the illuminated area on the mean surface, $\langle\cdot\rangle$ denotes the ensemble average over surface realizations, and $Z_0$ is the free-space wave impedance. $P_i$ denotes the incident power, and $dP_o/d\Omega_s$ denotes the scattered power per unit solid angle. Substituting expressions and the far-field scattered field into the BRDF definition yields
\begin{equation}
f_s=\frac{k^4}{\pi^2}\cos\theta_i\cos\theta_s|Q|^2\frac{\left\langle|H(\mathbf q_s-\mathbf q_i)|^2\right\rangle}{A}.
\end{equation}

We characterize the statistical properties of surface height using the 2D power spectral density (PSD) $S_h(f_x,f_y)$, assuming an isotropic Gaussian surface-height autocovariance parameterized by the root-mean-square (RMS) roughness $\sigma$ and correlation length $l$. The complete BRDF is then expressed as:
\begin{align}
f_{r}(\boldsymbol{\omega}_i, &\boldsymbol{\omega}_o)=\underbrace{\frac{16\pi^2}{\lambda^4}(\mathbf{n}\cdot\boldsymbol{\omega}_i)(\mathbf{n}\cdot\boldsymbol{\omega}_o)|Q|^2S_h(f_x,f_y)}_{\text{scattering}}\notag\\
&+\underbrace{\frac{(1-m)a}{\pi}}_{\text{diffuse}}+\underbrace{F(\boldsymbol{\omega}_i,a,m)\delta_{\mathbb{S}^2}(\boldsymbol{\omega}_o-\boldsymbol{\omega}_r)}_{\text{specular}},\label{eq:brdf}
\end{align}
where $F(\boldsymbol{\omega},a,m)$ denotes the Schlick approximation of the Fresnel reflectance, $a$ is the diffuse albedo, and $m$ is the metallic value. $\boldsymbol{\omega}_r=2(\mathbf n\cdot\boldsymbol{\omega}_i)\mathbf n-\boldsymbol{\omega}_i$ is the reflected direction. The three terms represent roughness-induced scattering, effective diffuse reflection associated with volume scattering, and coherent specular reflection from the mean surface, respectively. The surface-height PSD is given by $S_h(f_x,f_y)=\pi\sigma^2l^2\exp[-\left(\frac{\pi l f}{\lambda}\right)^2]$, where $f=\|(\mathbf I-\mathbf n\mathbf n^\top)(\boldsymbol{\omega}_i+\boldsymbol{\omega}_o)\|$. Since the four polarization channels exhibit different scattering responses, we average their contributions to obtain an effective polarization factor $\bar Q$. We then apply the Butler approximation~\cite{Butler:15} to obtain the following closed-form expression:
\begin{equation}
|\bar Q|^2=\frac{4u^4}{\cos\theta_i\cos\theta_s(\cos\theta_i+\cos\theta_s)^2}F(u,a,m),
\end{equation}
where $u^2 = \frac12(1+\cos\theta_i\cos\theta_s-\sin\theta_i\sin\theta_s\cos\phi_s)$. Further details are provided in Appendix~\ref{app:spm}.

\subsection{2DGS in Radio Fields}
To represent scene geometry for electromagnetic wave transport, we adopt 2DGS~\cite{10.1145/3799902.3811054, Huang2DGS2024}, which provides explicit surface normals and depth information. Each 2D Gaussian is parameterized by its center position $\mathbf p$, in-plane scales $s_u$ and $s_v$, orientation $\mathbf q$, and opacity $\alpha$. Following the BRDF formulation in Eq.~\ref{eq:brdf}, we further associate each Gaussian with four electromagnetic parameters: the metallic parameter $m\in[0,1]$, diffuse albedo $a\in[0,1]$, RMS roughness $\sigma\in[0,\frac{0.3}{k}]$, and correlation length $l\in[0,\frac6k]$.

\noindent\textbf{2D Gaussian Ray Tracing.} 
Given a ray with origin $\mathbf O$ and direction $\boldsymbol\omega$, an intersection point with the Gaussian is parameterized as $\mathbf x=\mathbf O+d\boldsymbol\omega$, where $d$ denotes the distance traveled along the ray. The spatial response of the Gaussian at this point is defined as
\begin{equation}
g=\exp\left[\frac{-\left((\mathbf x-\mathbf p)\cdot\mathbf t_{u}\right)^2}{2s_{u}^2}-\frac{\left((\mathbf x-\mathbf p)\cdot\mathbf t_{v}\right)^2}{2s_{v}^2}\right],
\end{equation}
where $\mathbf t_{u}$ and $\mathbf t_{v}$ are two principal tangential vectors. For a scene represented by a set of 2D Gaussians, the ray is traced forward and sequentially intersects $N_g$ Gaussians. At each intersection, the Gaussian quantity $m_i\in \{\mathbf p, s_u, s_v, \mathbf q, \alpha, m, a, \sigma, l\}$ and ray distance $d_i$ are accumulated using alpha compositing to obtain the macro-level quantity $M$ and distance $D$:
\begin{equation}
M=\sum_i^{N_g} w_i m_i,\quad T_i=\prod_{j=1}^{i}(1-\alpha_j g_j),
\end{equation}
where $w_i=T_{i-1}\alpha_i g_i$ is the blending weight of the $i$-th intersection. The macro-level intersection point is computed as $\mathbf{x}=\mathbf{O}+\widetilde D\boldsymbol{\omega}$, where $\widetilde D=\frac{D}{1-T_{N_g}}$. A new ray is then traced from $\mathbf{x}$, accumulating attributes $M'$ and yielding the next intersection point $\mathbf{x}'=\mathbf{x}+\widetilde D'\boldsymbol{\omega}'$. This process repeats until the path terminates or the bounce limit is reached.

\noindent\textbf{Geometry Reconstruction.}
Given multi-view images and an initial coarse RGB point cloud, we represent the scene as a collection of 2D Gaussians. For radiance supervision, following prior work on radiant scene reconstruction~\cite{Huang2DGS2024,10.1145/3592433,10.1145/3687934}, we employ the standard color reconstruction loss $\mathcal{L}_c$. For geometry supervision, we align the rendered normals $\tilde{\mathbf N}$ and normalized depth gradients $\tilde{\mathbf N}_d$ with monocular normal maps $\mathbf N$ estimated by StableNormal~\cite{10.1145/3687971}: $\mathcal{L}_n=\|\mathbf{1}-\tilde{\mathbf N}\cdot\mathbf N\|_1$, $\mathcal{L}_d=\|\mathbf{1}-\tilde{\mathbf N}_d\cdot\mathbf N\|_1$. The overall objective is defined as
\begin{equation}
\mathcal{L}_g=\lambda_c\mathcal{L}_c+\lambda_n\mathcal{L}_n+\lambda_d\mathcal{L}_d.
\end{equation}
After scene reconstruction, we freeze the geometric parameters $\{\mathbf p, s_u, s_v, \mathbf q, \alpha\}$ of all Gaussians and optimize the four electromagnetic parameters $\{m, a, \sigma, l\}$ within the subsequent RF relighting framework.

\subsection{Point-Source Two-Bounce Framework}
In RF relighting, we model the transmitter as a point source under the assumption that the scene lies in the far-field region of the antenna. Under this approximation, direct illumination reaches each surface point along a single geometrically determined direction. Based on this model, we formulate an RF relighting framework that explicitly accounts for direct illumination from the transmitter and interactions between surfaces. We first describe the direct illumination incident on a surface point from the transmitter.

\noindent\textbf{Direct incidence.}
Consider a transmitter located at $\mathbf{x}_t$ that radiates radio waves into the surrounding environment. The incident power density at a surface point $\mathbf{x}$ is modeled as
\begin{equation}
S_{t}(\mathbf{x})=\frac{P_t G_t(-\boldsymbol{\omega}_{t})}{4\pi \|\mathbf{x}-\mathbf{x}_t\|^2}T(\mathbf{x}, \mathbf{x}_t),
\end{equation}
where $P_t$ and $G_t(-\boldsymbol{\omega}_{t})$ denote the transmitted power and the antenna gain in the propagation direction, respectively. The unit vector $\boldsymbol{\omega}_t=\frac{\mathbf{x}_t-\mathbf{x}}{\|\mathbf{x}_t-\mathbf{x}\|}$ points from the surface point toward the transmitter. The distance factor accounts for free-space geometric spreading. Since radio waves can penetrate certain building materials, we do not impose hard geometric occlusion and allow signals to reach surface points beyond intervening objects. The attenuation along the path is modeled by the cumulative Gaussian transmittance $T(\mathbf{x},\mathbf{x}_t)$, defined as $T(\mathbf{x}, \mathbf{x}_t)=T_{i-1}$, where $i$ denotes the number of Gaussians along the path.

\noindent\textbf{One Bounce} is important for realistic rendering, particularly in areas occluded by other objects or surfaces. For an incident wave arriving at $\mathbf{x}$ and scattered in the direction $\boldsymbol{\omega}_o$, the scattered power per unit solid angle is given by
\begin{equation}
S_{o}^{(1)}(\boldsymbol{\omega}_o, \mathbf{x})=f_{r}(\boldsymbol{\omega}_{t}, \boldsymbol{\omega}_o)S_{t}(\mathbf{x})(\mathbf{n}\cdot\boldsymbol{\omega}_t)A(\mathbf{x}),
\end{equation}
where $A(\mathbf{x})=2\pi(\mathbf{n}\cdot\boldsymbol{\omega}_o)\sum_{i=1}^{N_g}w_i s_{u,i}s_{v,i}$ is the projected area of the surface patch illuminated by the outgoing wave.

This formulation reveals a limitation of one-bounce modeling: the incident direction $\boldsymbol{\omega}_t$ at each surface point is uniquely determined by the transmitter location. Consequently, the model learns the scattering response only under this incident direction, rather than capturing its dependence on general incident-outgoing direction pairs. When the transmitter is moved, the incident direction changes, introducing a distribution shift that can impair generalization. To mitigate this limitation, we explicitly incorporate a two-bounce framework, allowing waves reflected from other surfaces to illuminate a surface from additional incident directions. This introduces greater diversity in the direction pairs, providing more comprehensive constraints on the surface scattering response and thereby improving generalization to unseen transmitter locations.

\noindent\textbf{Two Bounce} refers to a process in which a wave emitted by a transmitter is first scattered at $\mathbf{x}_1$ and subsequently reaches $\mathbf{x}_2$, where it undergoes a second scattering event. The resulting incident power density at $\mathbf{x}_2$ from direction $\boldsymbol{\omega}_i$ is given by
\begin{equation}
S_i^{(2)}(\boldsymbol{\omega}_i, \mathbf{x}_2)=\frac{T(\mathbf{x}_1, \mathbf{x}_2)}{\|\mathbf{x}_2-\mathbf{x}_1\|^2}S_o^{(1)}(-\boldsymbol{\omega}_i, \mathbf{x}_1).
\end{equation}
The scattered power per unit solid angle toward the receiver is obtained by integrating the contributions from all incident directions:
\begin{align}
&S_{o}^{(2)}(\boldsymbol{\omega}_r, \mathbf{x}_2)\notag\\
&=\int_{\Omega}f_{r}(\boldsymbol{\omega}_i, \boldsymbol{\omega}_r)S_i^{(2)}(\boldsymbol{\omega}_i, \mathbf{x}_2)(\mathbf{n}_2\cdot\boldsymbol{\omega}_i)A(\mathbf{x}_2)d\boldsymbol{\omega}_i,\label{eq:two_bounce}
\end{align}
where $\boldsymbol{\omega}_r=\frac{\mathbf{x}_r-\mathbf{x}_2}{\|\mathbf{x}_r-\mathbf{x}_2\|}$ denotes the unit direction from $\mathbf{x}_2$ to the receiver $\mathbf{x}_r$, and $\mathbf n_2$ is the surface normal at $\mathbf{x}_2$.

\begin{table*}[t]
    \centering
    \caption{Quantitative comparison with SOTA methods on the NIST and WiSegRT datasets. RF-3DGS and GSRF are evaluated only in the few-shot setting because they do not support zero-shot evaluation. \textbf{Bold} and \underline{underline} denote the best and the second-best results.}
    \label{tab:main_results}
    \small
    \renewcommand{\arraystretch}{0.9}
    \begin{tabularx}{\textwidth}{
    >{\raggedright\arraybackslash}p{1.9cm}
    >{\raggedright\arraybackslash}p{2.2cm}
    *{9}{>{\centering\arraybackslash}X}
    >{\centering\arraybackslash}p{0.9cm}
    }
    \toprule
    \multirow{2}{*}{Dataset} &
    \multirow{2}{*}{Methods} &
    \multicolumn{3}{c}{NRS} &
    \multicolumn{3}{c}{Relighting (Zero-Shot)} &
    \multicolumn{3}{c}{Relighting (Few-Shot)} &
    \multirow{2}{*}{\shortstack{Time\\(hours)}} \\
    \cmidrule(lr){3-5}
    \cmidrule(lr){6-8}
    \cmidrule(lr){9-11}
    & &
    MAE$\downarrow$ & PSNR$\uparrow$ & SSIM$\uparrow$ &
    MAE$\downarrow$ & PSNR$\uparrow$ & SSIM$\uparrow$ &
    MAE$\downarrow$ & PSNR$\uparrow$ & SSIM$\uparrow$ &
    \\
    \midrule

    \multirow{6}{*}{\shortstack{NIST\\(60 GHz)}}
    & RF-3DGS~\cite{11355734}
    & \underline{6.448} & 17.14 & 0.643
    & -- & -- & --
    & 12.625 & 9.95 & 0.472
    & \textbf{0.03} \\

    & GSRF~\cite{NEURIPS2025_5b5618e7}
    & 7.101 & 18.18 & 0.536
    & -- & -- & --
    & 12.442 & \underline{14.95} & \underline{0.629}
    & 0.61 \\

    & BiWGS~\cite{11720435}
    & 8.649 & 16.88 & 0.646
    & 24.584 & 9.35 & \underline{0.612}
    & 21.623 & 10.44 & \underline{0.641}
    & 2.09 \\

    & RFIR~\cite{wang2026rfir}
    & 6.983 & \underline{18.35} & \underline{0.666}
    & 18.327 & \underline{11.60} & 0.592
    & \underline{12.117} & 14.94 & 0.624
    & 1.45 \\

    & URF-GS~\cite{wen2026urfgs}
    & 10.655 & 14.71 & 0.598
    & \underline{16.972} & 9.49 & 0.488
    & 15.331 & 11.21 & 0.535
    & \underline{0.18} \\

    & \textbf{WRGS (ours)}
    & \textbf{5.423} & \textbf{20.04} & \textbf{0.712}
    & \textbf{8.047} & \textbf{17.54} & \textbf{0.694}
    & \textbf{7.848} & \textbf{18.34} & \textbf{0.731}
    & 1.14 \\

    \midrule

    \multirow{6}{*}{\shortstack{WiSegRT\\(2.4 GHz)}}
    & RF-3DGS~\cite{11355734}
    & \underline{22.275} & 13.44 & 0.591
    & -- & -- & --
    & 45.705 & 7.79 & 0.431
    & \textbf{0.03} \\

    & GSRF~\cite{NEURIPS2025_5b5618e7}
    & 24.123 & 12.82 & \underline{0.660}
    & -- & -- & --
    & 43.564 & 9.33 & 0.458
    & 0.55 \\

    & BiWGS~\cite{11720435}
    & 26.084 & 13.23 & 0.631
    & 28.305 & 10.86 & \underline{0.517}
    & 25.496 & 11.46 & \underline{0.532}
    & 1.01 \\

    & RFIR~\cite{wang2026rfir}
    & 23.505 & \underline{14.32} & 0.629
    & \underline{25.845} & \underline{11.91} & 0.493
    & 23.115 & \underline{12.36} & 0.516
    & 0.96 \\

    & URF-GS~\cite{wen2026urfgs}
    & 25.648 & 12.79 & 0.573
    & 33.840 & 9.36 & 0.378
    & \underline{21.957} & 12.33 & 0.488
    & \underline{0.14} \\

    & \textbf{WRGS (ours)}
    & \textbf{15.257} & \textbf{16.49} & \textbf{0.705}
    & \textbf{18.905} & \textbf{13.23} & \textbf{0.531}
    & \textbf{17.059} & \textbf{14.01} & \textbf{0.563}
    & 0.88 \\

    \bottomrule
    \end{tabularx}
\end{table*}

\noindent\textbf{Global Illumination.}
The received power along each direction $\boldsymbol{\omega}_r$ consists of contributions from one-bounce and two-bounce reflections. A direct-path contribution is additionally included when the transmitter-to-receiver direction coincides with $\boldsymbol{\omega}_r$. The received power is expressed as
\begin{align}
S_r(\boldsymbol{\omega}_r)&=\left[S_o^{(1)}(\boldsymbol{\omega}_r, \mathbf{x}_2)+S_o^{(2)}(\boldsymbol{\omega}_r, \mathbf{x}_2)\right]\frac{A_r T(\mathbf x_2,\mathbf x_r)}{\|\mathbf{x}_r-\mathbf{x}_2\|^2}\notag\\
&+A_r S_t(\mathbf x_r)\delta(\boldsymbol{\omega}_r-\frac{\mathbf{x}_r-\mathbf{x}_t}{\|\mathbf{x}_r-\mathbf{x}_t\|}),
\end{align}
where $A_r=\frac{\lambda^2G_r(\boldsymbol{\omega}_r)}{4\pi}$ denotes the effective aperture of the receiver. In practice, we approximate the angular integral using uniform sampling of incident directions over $\Omega$.

In practical antenna measurements, finite angular resolution causes signals arriving from a given direction to spread across neighboring directions~\cite{9298913}. To account for this effect, we convolve the predicted APS with a Gaussian kernel before comparing it with the ground truth.

Furthermore, the large dynamic range of the APS makes reconstruction losses in the linear domain less effective, as high-power directions can dominate the optimization. Therefore, we transform both the predicted and ground-truth APS into the dB domain before computing the reconstruction loss. This logarithmic transformation compresses the dynamic range, providing more balanced supervision across directions with different power levels. We define the final objective by combining L1 and SSIM losses:
\begin{equation}
    \mathcal{L}_e=\lambda_e \mathcal{L}_1+(1-\lambda_e)\mathcal{L}_{\text{SSIM}}.
\end{equation}

%% file: sec/4_experiment.tex
\section{Experiment}
\subsection{Experimental Setup}
\noindent\textbf{Datasets and Metrics.}
We evaluate our method on two propagation scenarios, WiSegRT~\cite{10556262} and NIST~\cite{10501217}. WiSegRT comprises ten indoor scenes with object-level material segmentation at 2.4 GHz, while the NIST scenario features a 14 $\times$ 15 m$^2$ lobby at 60 GHz. Dense sampling of receiver positions is required for real-world measurements, while sequential acquisition can introduce temporal channel variations, leading to distribution shifts across measurements. Simultaneous acquisition, in contrast, requires complex synchronization and calibration and may introduce antenna mutual coupling. We therefore generate propagation data from the 3D scenes using Sionna RT~\cite{10465179}. We evaluate reconstruction quality using MAE, PSNR, and SSIM. During evaluation, MAE is computed in the original dB scale, whereas PSNR and SSIM are computed on the normalized data. Further details are provided in Appendix~\ref{app:data}.

\noindent\textbf{Baselines and Implementation Details.}
We benchmark our method against SOTA methods for RF relighting, including BiWGS~\cite{11720435}, RFIR~\cite{wang2026rfir} and URF-GS~\cite{wen2026urfgs}. For NRS, we compare against RF-3DGS~\cite{11355734} and GSRF~\cite{NEURIPS2025_5b5618e7}. For each Tx position, we generate 3,200 samples, each containing the 3D coordinates of a randomly selected Rx and its corresponding APS. We use eight uniformly sampled incident directions to approximate the angular integral. Our evaluation covers three settings:
\begin{itemize}
    \item \textbf{NRS}: Samples at each Tx position are randomly split into 80\% for training and 20\% for testing.
    \item \textbf{Few-Shot RF Relighting}: 10 samples from the unseen Tx position are added to the training set, and the model is evaluated on the remaining held-out samples.
    \item \textbf{Zero-Shot RF Relighting}: No samples from the unseen Tx position are used for training, and the model is directly evaluated on the corresponding test set.
\end{itemize}

\begin{table*}[t]
    \centering
    \caption{Quantitative comparison of zero-shot RF relighting performance across different wall materials on the NIST dataset. Since walls occupy the largest area and strongly affect electromagnetic propagation, we vary only the wall material while keeping all walls uniform and other objects unchanged.}
    \label{tab:diff_materials}
    \small
    \renewcommand{\arraystretch}{0.9}
    \begin{tabularx}{\textwidth}{>{\raggedright\arraybackslash}p{2cm}*{12}{C}}
    \toprule
    \multirow{2}{*}{Methods}
    & \multicolumn{3}{c}{Gypsum}
    & \multicolumn{3}{c}{Concrete}
    & \multicolumn{3}{c}{Glass}
    & \multicolumn{3}{c}{Wood} \\
    \cmidrule(lr){2-4}
    \cmidrule(lr){5-7}
    \cmidrule(lr){8-10}
    \cmidrule(lr){11-13}
    & MAE$\downarrow$ & PSNR$\uparrow$ & SSIM$\uparrow$
    & MAE$\downarrow$ & PSNR$\uparrow$ & SSIM$\uparrow$
    & MAE$\downarrow$ & PSNR$\uparrow$ & SSIM$\uparrow$
    & MAE$\downarrow$ & PSNR$\uparrow$ & SSIM$\uparrow$ \\
    \midrule

    BiWGS~\cite{11720435}
    & 24.584 & 9.35 & \underline{0.612}
    & 24.193 & 10.27 & 0.604
    & 24.218 & 10.26 & 0.603
    & 24.246 & 10.25 & 0.604 \\

    RFIR~\cite{wang2026rfir}
    & 18.327 & \underline{11.60} & 0.592
    & 31.522 & 8.19 & \underline{0.612}
    & 31.493 & 8.21 & \underline{0.604}
    & 31.462 & 8.20 & \underline{0.611} \\

    URF-GS~\cite{wen2026urfgs}
    & \underline{16.972} & 9.49 & 0.488
    & \underline{15.037} & \underline{11.72} & 0.552
    & \underline{15.074} & \underline{11.73} & 0.551
    & \underline{14.995} & \underline{11.74} & 0.533 \\

    \textbf{WRGS (ours)}
    & \textbf{8.047} & \textbf{17.54} & \textbf{0.694}
    & \textbf{7.734} & \textbf{18.13} & \textbf{0.686}
    & \textbf{7.862} & \textbf{17.97} & \textbf{0.666}
    & \textbf{7.852} & \textbf{17.98} & \textbf{0.667} \\

    \bottomrule
    \end{tabularx}
\end{table*}

\begin{figure*}[t]
    \centering
    \includegraphics[width=\textwidth]{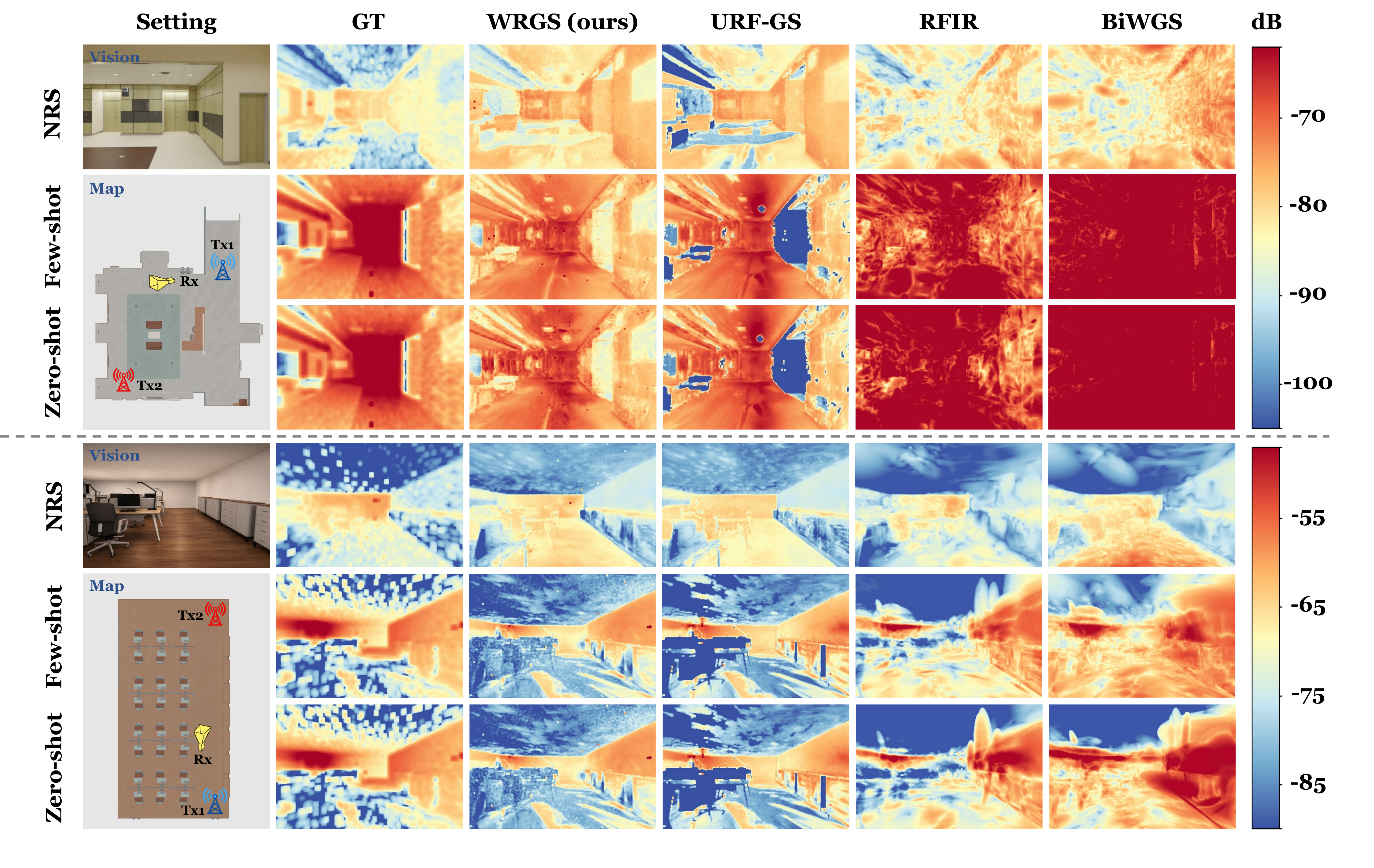}
    \caption{Visual comparisons of WRGS with SOTA methods on NRS and RF relighting. For each scene, the leftmost column (Setting) shows the receiver-view image and the corresponding floor plan, with the locations of Tx and the Rx indicated. NRS is evaluated using Tx1, whereas RF relighting is evaluated using Tx2.}
    \label{fig:visual}
\end{figure*}

\subsection{Main Results}

\noindent\textbf{Quantitative Comparison with SOTA Methods.}
Tab.~\ref{tab:main_results} presents quantitative comparisons with SOTA methods on NIST (60 GHz) and WiSegRT (2.4 GHz). Our method achieves the best performance across all metrics and evaluation settings on both datasets. Notably, for RF relighting, WRGS achieves significant improvements in both zero-shot and few-shot settings, reducing MAE by 52.6\% and 26.9\% in zero-shot, and by 35.2\% and 22.3\% in few-shot, respectively. These results demonstrate the effectiveness of our method in synthesizing RF fields at unseen Tx positions across different frequency bands and propagation scenarios.

\noindent\textbf{Runtime Comparison.}
Our method requires 1.14 hours on NIST and 0.88 hours on WiSegRT. Although WRGS has a higher runtime than RF-3DGS, GSRF, and URF-GS, it remains faster than BiWGS and RFIR on both datasets. Considering the substantial improvements in reconstruction and relighting quality, WRGS achieves a favorable trade-off between performance and computational cost.

\noindent\textbf{Robustness to Different Surface Materials.}
Tab.~\ref{tab:diff_materials} compares zero-shot RF relighting performance across four wall materials on the NIST dataset. Our method consistently achieves the best results across gypsum, concrete, glass, and wood. Moreover, our method maintains stable performance across different materials, with only minor variations in all three metrics. These results demonstrate the robustness of our method across different surface materials.

\noindent\textbf{Visual Comparison.}
Fig.~\ref{fig:visual} visualizes the ground-truth (GT) and predicted APS maps. For NRS, WRGS predictions closely match the GT in spatial patterns and signal distributions. In comparison, URF-GS produces prominent artifacts and distorted regions, whereas RFIR and BiWGS fail to recover clear spatial patterns. For RF relighting, WRGS consistently preserves major propagation structures, under both few-shot and zero-shot settings, while competing methods exhibit pronounced artifacts and structural distortions. These results demonstrate the ability of WRGS to generalize to unseen Tx positions, even in the zero-shot setting without target-Tx observations.

\subsection{Ablation Study}

\begin{table}[t]
    \centering
    \caption{Ablation study on the number of Tx positions and bounces under a fixed data budget. All configurations employ the SPM-based BRDF.}
    \label{tab:diff_tb}
    \small
    \renewcommand{\arraystretch}{0.9}
    \begin{tabularx}{\columnwidth}{l*{6}{>{\centering\arraybackslash}X}}
    \toprule
    & \multicolumn{3}{c}{NRS}
    & \multicolumn{3}{c}{Relighting (Zero-Shot)} \\
    \cmidrule(lr){2-4}
    \cmidrule(lr){5-7}
    & MAE$\downarrow$ & PSNR$\uparrow$ & SSIM$\uparrow$
    & MAE$\downarrow$ & PSNR$\uparrow$ & SSIM$\uparrow$ \\
    \midrule
    1T-1B
    & 7.719 & 17.04 & 0.626
    & 19.481 & 10.65 & 0.480 \\

    1T-2B
    & \underline{5.423} & \underline{20.04} & \underline{0.712}
    & \underline{8.047} & \underline{17.54} & 0.694 \\

    3T-1B
    & 8.304 & 17.03 & 0.538
    & 8.696 & 16.75 & \underline{0.699} \\

    3T-2B
    & \textbf{4.251} & \textbf{23.54} & \textbf{0.826}
    & \textbf{5.152} & \textbf{22.01} & \textbf{0.784} \\

    \bottomrule
    \end{tabularx}
\end{table}

\noindent\textbf{Effect of Two-Bounce Framework.}
Tab.~\ref{tab:diff_tb} compares four configurations with different numbers of Tx positions and maximum bounce orders under a fixed data budget. In the notation $n$T-$m$B, $n$ denotes the number of training Tx positions, with the total training samples distributed equally among these positions, and $m$ denotes the maximum number of reflections modeled by the propagation framework. Introducing the two-bounce framework consistently improves all metrics in both NRS and RF relighting under 1T and 3T configurations. These results demonstrate the effectiveness of explicitly modeling inter-surface reflections.

We further investigate the effect of Tx diversity. For zero-shot relighting, increasing the number of Tx positions from one to three under 1B substantially reduces MAE from 19.48 to 8.696. This improvement is consistent with our motivation: for a fixed Tx, each surface point receives signals from only one direct incident direction, limiting the diversity of observed surface responses. However, even with three Tx positions, 3T-1B remains inferior to 3T-2B across all metrics in both NRS and RF relighting. This further highlights the benefit of the two-bounce framework, which introduces indirect propagation paths and enriches incident-outgoing direction combinations beyond what can be achieved by increasing Tx diversity alone.

\noindent\textbf{Effect of BRDF.}
Tab.~\ref{tab:brdf} compares four BRDF formulations and evaluates the contributions of the diffuse and specular components by removing each component in turn. Our SPM-based BRDF achieves the best performance across all metrics in both NRS and zero-shot RF relighting. Among the baselines, RER provides a scattering lobe better suited to electromagnetic scattering, but its empirical formulation limits generalization. CT is less suitable for RF scattering due to its lack of explicit electromagnetic wave modeling. BSH exhibits the largest performance degradation from NRS to RF relighting, indicating limited generalization of its data-driven angular response. The ablation results show that diffuse scattering has a greater impact on performance due to its broad angular coverage, while specular reflection has a smaller contribution. These findings highlight the importance of modeling electromagnetic wave effects.

Additional ablation studies on the number of sampled rays, Gaussian convolution, and supervision in the dB domain are provided in Appendices~\ref{app:sample},~\ref{app:gaussian}, and~\ref{app:db}, respectively.

\begin{table}[t]
    \centering
    \caption{Ablation study of BRDF models within the same two-bounce framework.}
    \label{tab:brdf}
    \small
    \renewcommand{\arraystretch}{0.9}
    \begin{tabularx}{\columnwidth}{l*{6}{>{\centering\arraybackslash}X}}
    \toprule
    & \multicolumn{3}{c}{NRS}
    & \multicolumn{3}{c}{Relighting (Zero-Shot)} \\
    \cmidrule(lr){2-4}
    \cmidrule(lr){5-7}
    & MAE$\downarrow$ & PSNR$\uparrow$ & SSIM$\uparrow$
    & MAE$\downarrow$ & PSNR$\uparrow$ & SSIM$\uparrow$ \\
    \midrule
    CT~\cite{10.1145/357290.357293}
    & 6.312 & 18.90 & \underline{0.707}
    & 10.482 & 16.33 & 0.630 \\

    RER~\cite{10137406}
    & 5.903 & 19.66 & 0.683
    & 10.500 & 16.24 & 0.571 \\

    BSH~\cite{11125006}
    & 6.391 & 19.24 & 0.703
    & 10.656 & 16.19 & 0.558 \\

    w/o diff
    & 5.552 & 19.98 & 0.681
    & 8.442 & 17.46 & 0.667 \\

    w/o spec
    & \underline{5.519} & \underline{20.02} & 0.705
    & \underline{8.052} & \underline{17.53} & \underline{0.687} \\

    SPM (ours)
    & \textbf{5.423} & \textbf{20.04} & \textbf{0.712}
    & \textbf{8.047} & \textbf{17.54} & \textbf{0.694} \\

    \bottomrule
    \end{tabularx}
\end{table}

\subsection{Data Efficiency Analysis}
\begin{figure}[t]
    \centering
    \includegraphics[width=\columnwidth]{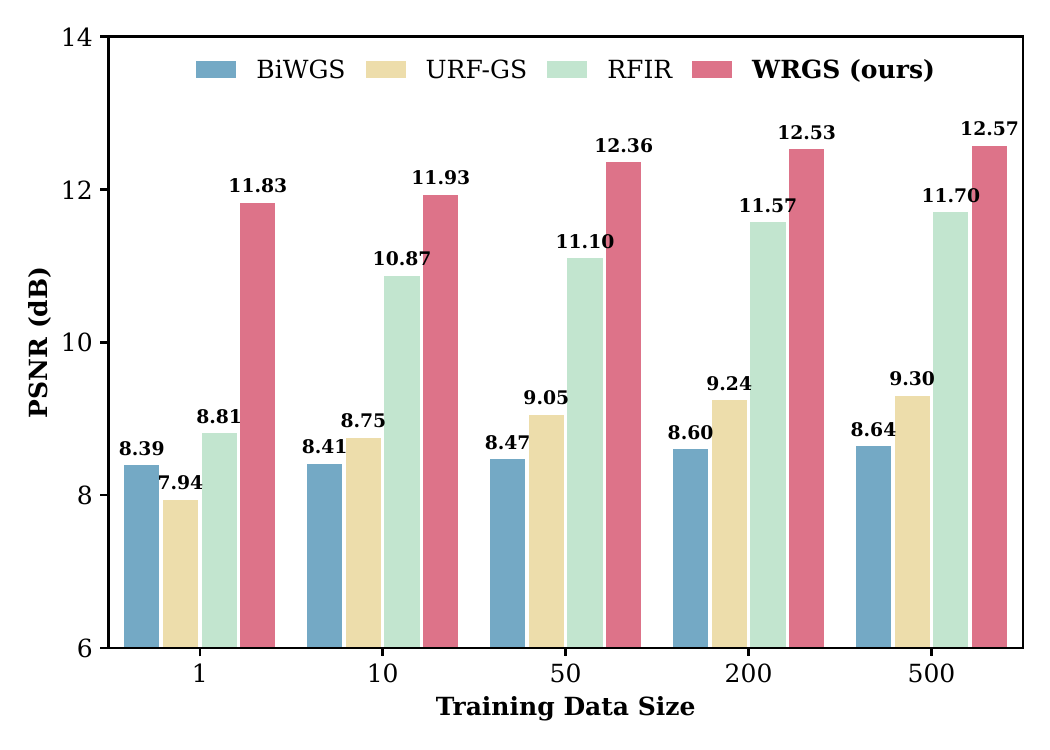}
    \caption{Performance comparison of zero-shot RF relighting across different training data sizes.}
    \label{fig:ds}
\end{figure}

Fig.~\ref{fig:ds} illustrates the effect of training data size on zero-shot RF relighting. Our method consistently outperforms all baselines across all training data sizes, demonstrating strong data efficiency. With only one training sample, our method achieves a PSNR of 11.83 dB, surpassing RFIR trained with 500 samples. These results demonstrate that our method maintains strong relighting performance with substantially fewer training samples.


%% file: sec/5_conclusion.tex
\section{Conclusion}
In this paper, we present WRGS, a physically grounded framework for RF relighting that explicitly models electromagnetic scattering and inter-surface propagation. Specifically, we derive an SPM-based BRDF to capture wave scattering from rough surfaces and introduce a 2DGS-based two-bounce Framework to enrich incident-outgoing direction pairs under fixed Tx configurations. Experiments on NIST and WiSegRT demonstrate SOTA performance in NRS and zero-shot/few-shot RF relighting, validating the effectiveness of both proposed components.

\section{LLM Usage Statement}
ChatGPT was used to polish the language throughout the manuscript and generate selected visual elements in Fig.~\ref{fig:task} and Fig.~\ref{fig:architecture}. The research ideas and methodological and experimental designs were developed by the authors. The authors independently performed all data generation, experiment execution, quantitative evaluations, and result analyses. All reported experimental results were obtained and verified by the authors.

%% file: sec/X_suppl.tex
\clearpage
\setcounter{page}{1}
\maketitlesupplementary

\appendix
\section{SPM-Based BRDF Formulation}\label{app:spm}
In the small perturbation method (SPM), electromagnetic scattering from a rough surface is analyzed by treating the surface height as a perturbation about its mean plane. The upper and lower half-spaces are occupied by air and a nonmagnetic dielectric, respectively, with relative permittivities $\epsilon_r^+=1$ and $\epsilon_r^-=\epsilon$. Both media are assumed to have a relative permeability of unity. Let $\mathbf E^\pm$ and $\mathbf M^\pm$ denote the total electric and normalized magnetic fields in the upper ($+$) and lower ($-$) half-spaces, respectively.

The total electric and normalized magnetic fields are expanded as perturbation series in the surface height:
\begin{align}
\mathbf E^\pm&=\mathbf E^{\pm(0)}+\mathbf E^{\pm(1)}+O(h^2),\\
\mathbf M^\pm&=\mathbf M^{\pm(0)}+\mathbf M^{\pm(1)}+O(h^2),
\end{align}
where the superscripts $(0)$ and $(1)$ denote the zeroth- and first-order contributions, respectively. The zeroth-order fields correspond to the solution for a flat interface, whereas the first-order fields describe the leading-order corrections induced by surface roughness. The jump of a field quantity $F$ across the interface is defined as $[F]=F^+-F^-$.

\subsection{Coordinate System}
We establish a fixed Cartesian coordinate system with the mean interface at $z=0$, the positive $z$-axis pointing toward the air region, and the rough interface described by $z=h(\boldsymbol\rho)$. Let $\mathbf x$, $\mathbf y$, and $\mathbf z$ denote the Cartesian unit vectors. The in-plane position vector and gradient operator are defined as $\boldsymbol\rho=x\mathbf x+y\mathbf y$ and $\nabla_\parallel=\mathbf x\partial_x+\mathbf y\partial_y$, respectively. The incident and scattered unit propagation vectors, specified by their zenith and azimuth angles $(\theta_i,\phi_i)$ and $(\theta_s,\phi_s)$, respectively, are given by
\begin{align}
\mathbf k_i&=\sin\theta_i\cos\phi_i\mathbf x+\sin\theta_i\sin\phi_i\mathbf y-\cos\theta_i\mathbf z,\\
\mathbf k_s&=\sin\theta_s\cos\phi_s\mathbf x+\sin\theta_s\sin\phi_s\mathbf y+\cos\theta_s\mathbf z.
\end{align}

For $m=i,s$, we define $\mathbf t_m=\cos\phi_m\mathbf x+\sin\phi_m\mathbf y$ as the unit vector along the projection of $\mathbf k_m$ onto the mean interface. The corresponding $s$- and $p$-polarization basis vectors are defined as $\mathbf s_m=\mathbf z\times\mathbf t_m$ and $\mathbf p_m=\mathbf k_m\times\mathbf s_m$, respectively. $\mathbf s_m$ is normal to the corresponding plane of incidence or scattering and lies in the mean interface, whereas $\mathbf p_m$ lies in that plane and is perpendicular to both $\mathbf k_m$ and $\mathbf s_m$.

Let $k=2\pi/\lambda$ denote the free-space wavenumber, where $\lambda$ is the wavelength. The transverse wave vectors of the incident and scattered waves are defined as $\mathbf K_i=k\sin\theta_i\mathbf t_i$ and $\mathbf K_s=k\sin\theta_s\mathbf t_s$, respectively. Their difference is denoted by $\Delta\mathbf K=\mathbf K_s-\mathbf K_i$, and the relative azimuth angle is defined as $\Delta\phi=\phi_s-\phi_i$.

\subsection{Far-Field Scattered Field}
In the homogeneous air region above the rough interface, the first-order scattered field satisfies the Helmholtz equation and can be expressed as a superposition of plane-wave components with different transverse wave vectors, subject to the outgoing-wave condition. Decomposing each component into the corresponding $s$- and $p$-polarized basis vectors yields the angular-spectrum representation
\begin{align}
\widetilde{\mathbf E}_{\alpha}^{(1),+}(\boldsymbol\rho,z)&=a_{\alpha s}(\mathbf K)\mathbf s_s(\mathbf K)+a_{\alpha p}(\mathbf K)\mathbf p_s(\mathbf K),\notag\\
\mathbf E_\alpha^{(1)}(\boldsymbol\rho,z)&=\int \frac{\widetilde{\mathbf E}_{\alpha}^{(1),+}}{(2\pi)^2}e^{j\mathbf K\cdot\boldsymbol\rho+j\alpha_0(\mathbf K)z}d^2\mathbf K,\label{E_total}
\end{align}
where $\mathbf e_\beta(\mathbf K)$ denotes the outgoing polarization basis vector, and $\alpha_0(\mathbf K)=\sqrt{k^2-|\mathbf K|^2}$ is the longitudinal wave number in air. The coefficient $a_{\alpha\beta}(\mathbf K)$ represents the angular-spectrum amplitude of the $\beta$-polarized component for an incident polarization $\alpha$, which is to be determined from the first-order electromagnetic boundary conditions.

To evaluate the scattered field in the far field, we consider an observation point at a distance $r$ from the origin in the direction specified by the scattering angles $(\theta_s,\phi_s)$. Its coordinates are given by $\boldsymbol\rho=r\sin\theta_s\mathbf t_s$ and $z=r\cos\theta_s$. Substituting these coordinates into the exponential term in (\ref{E_total}) gives $e^{jr\Psi(\mathbf K)}$, where the phase function is defined as
\begin{equation}
\Psi(\mathbf K)=\sin\theta_s\mathbf t_s\cdot\mathbf K+\cos\theta_s\alpha_0(\mathbf K).
\end{equation}
For sufficiently large $r$, the angular-spectrum integral can be evaluated asymptotically using the stationary-phase method. The dominant contribution arises from the stationary point satisfying $\nabla_{\mathbf K}\Psi=0$. Combining this condition with the dispersion relation $ |\mathbf K|^2+\alpha_0^2(\mathbf K)=k^2$, we obtain
\begin{equation}
\mathbf K_s=\frac{\sin\theta_s}{\cos\theta_s}\alpha_0(\mathbf K_s)\mathbf t_s=k\sin\theta_s\mathbf t_s,
\end{equation}
at which $\Psi(\mathbf K_s)=k$. Thus, the stationary point corresponds to the transverse wave vector of the outgoing wave in the observation direction. Applying the two-dimensional stationary-phase approximation at $\mathbf K_s$ and extracting the $\beta$-polarized component yields
\begin{equation}
E_{\alpha\beta}^{(1)}(r)\approx-\frac{jk \cos\theta_s}{2\pi r}a_{\alpha\beta}(\mathbf K_s)e^{jkr}.\label{26}
\end{equation}
The far-field scattered electric field is therefore determined by the angular-spectrum amplitude evaluated at the observation direction. The remaining task is to derive an explicit expression for $a_{\alpha\beta}(\mathbf K_s)$ by solving the first-order electromagnetic boundary conditions in the Fourier domain.

\subsection{Zeroth-Order Fields}
To derive the first-order scattering fields, we first determine the zeroth-order electromagnetic fields, which provide the terms in the first-order boundary conditions. We begin by solving the plane-wave reflection and transmission problem at this flat dielectric interface. Let $E_0$ denote the amplitude of the incident electric field. The incident field varies along the mean interface according to the spatial phase factor $e^{j\mathbf K_i\cdot\boldsymbol{\rho}}$, where $j=\sqrt{-1}$ is the imaginary unit.

Applying the continuity conditions for the tangential electric and magnetic fields at the flat interface yields the Fresnel reflection coefficients $r_s$ and $r_p$, and the corresponding transmission coefficients $t_s$ and $t_p$. Specifically, the continuity conditions for the \(s\)- and \(p\)-polarized incident waves are given by
\begin{align}
    (1+r_s)E_0\mathbf{s}_i&=t_sE_0\mathbf{s}_i,\\
    \cos\theta_i(1-r_s)E_0\mathbf{t}_i&=\sqrt{\epsilon-\sin^2\theta_i}t_sE_0\mathbf{t}_i,\\
    \cos\theta_i(1-r_p)E_0\mathbf{t}_i&=\frac{\sqrt{\epsilon-\sin^2\theta_i}}{\sqrt{\epsilon}}t_pE_0\mathbf{t}_i,\\
    -(1+r_p)E_0\mathbf{s}_i&=-\sqrt{\epsilon}t_pE_0\mathbf{s}_i.
\end{align}
The common spatial phase factor is omitted from these boundary conditions for brevity. Solving the resulting system yields the Fresnel reflection coefficients:
\begin{align}
    r_s&=\frac{\cos\theta_i-\sqrt{\epsilon-\sin^2\theta_i}}{\cos\theta_i+\sqrt{\epsilon-\sin^2\theta_i}},\\
    r_p&=\frac{\epsilon\cos\theta_i-\sqrt{\epsilon-\sin^2\theta_i}}{\epsilon\cos\theta_i+\sqrt{\epsilon-\sin^2\theta_i}}.
\end{align}
Using the corresponding transmission coefficients, we obtain the zeroth-order tangential electric fields at the interface and the zeroth-order normal electric field in the dielectric:
\begin{align}
    \mathbf E_{\parallel,s}^{(0)}&=\frac{2\cos\theta_i}{\cos\theta_i+\sqrt{\epsilon-\sin^2\theta_i}}E_0\mathbf{s}_ie^{j\mathbf K_i\cdot\boldsymbol{\rho}},\label{33}\\
    \mathbf E_{\parallel,p}^{(0)}&=\frac{2\cos\theta_i\sqrt{\epsilon-\sin^2\theta_i}}{\epsilon\cos\theta_i+\sqrt{\epsilon-\sin^2\theta_i}}E_0\mathbf{t}_ie^{j\mathbf{K}_i\cdot\boldsymbol{\rho}},\label{34}\\
    E_{z,p}^{-(0)}&=\frac{2\cos\theta_i\sin\theta_i}{\epsilon\cos\theta_i+\sqrt{\epsilon-\sin^2\theta_i}}E_0e^{j\mathbf{K}_i\cdot\boldsymbol{\rho}}.\label{35}
\end{align}
For a vector field $\mathbf F$, the subscripts $\parallel$ and $z$ denote its tangential and normal components with respect to the mean interface, respectively. Specifically, $F_z$ denotes the scalar normal component, whereas $\mathbf F_z=F_z\mathbf z$ denotes the corresponding normal vector component. Since an $s$-polarized wave has no electric-field component normal to the mean interface, $E_{z,s}^{-(0)}=0$. We define $\mathbf e_{\parallel,s}$, $\mathbf e_{\parallel,p}$ as the zeroth-order tangential electric-field vectors, and $e_{z,p}^{-}$, $e_{z,s}^{-}$ as the corresponding zeroth-order normal electric-field components on the dielectric side, with the common factor $E_0 e^{j\mathbf K_i\cdot\boldsymbol{\rho}}$ factored out.

\subsection{First-Order Fields}
To determine the first-order scattered fields, we impose the electromagnetic boundary conditions on the actual rough interface, defined by $z=h(x,y)$. Unlike a flat interface, the orientation of the local tangent plane of a rough surface varies with the surface gradient $\nabla_\parallel h$. Consequently, the tangential continuity conditions involve both the tangential and normal components of the fields. The boundary conditions on the actual rough interface are therefore given by
\begin{equation}
    [\mathbf E_\parallel(x,y,h)]+[E_z(x,y,h)]\nabla_\parallel h=0.\label{eq:e1}
\end{equation}
Using the field perturbation expansion introduced above, we perform a Taylor expansion about the mean plane $z=0$. Retaining terms up to first order in the surface height gives
\begin{align}
    \mathbf E_\parallel(x,y,h)&=\mathbf E_\parallel^{(0)}(x,y,0)+\mathbf E_\parallel^{(1)}(x,y,0)\notag\\
    &+h\partial_z\mathbf E_\parallel^{(0)}(x,y,0)+O(h^2),\\
    E_z(x,y,h)&=E_z(x,y,0)+O(h).
\end{align}
Substituting these expansions into the electric-field boundary condition (\ref{eq:e1}) and retaining the first-order terms yields
\begin{equation}
    [\mathbf E_\parallel^{(0)}]+[\mathbf E_\parallel^{(1)}]+h[\partial_z\mathbf E_\parallel^{(0)}]+[E_z^{(0)}]\nabla_\parallel h=0.
\end{equation}
Since the zeroth-order tangential fields satisfy the flat-interface boundary conditions, $[\mathbf M_\parallel^{(0)}]=[\mathbf E_\parallel^{(0)}]=0$. Applying the same perturbation expansion to the normalized magnetic field $\mathbf M$ gives the corresponding first-order magnetic-field boundary condition. Combining the electric- and magnetic-field boundary conditions, we obtain
\begin{align}
    [\mathbf E_\parallel^{(1)}]&=-h[\partial_z\mathbf E_\parallel^{(0)}]-[E_z^{(0)}]\nabla_\parallel h,\label{eq:e2}\\
    [\mathbf M_\parallel^{(1)}]&=-h[\partial_z\mathbf M_\parallel^{(0)}]-[M_z^{(0)}]\nabla_\parallel h.\label{eq:m2}
\end{align}

To simplify the first-order boundary conditions, we use the Maxwell equations in the two homogeneous media. The normalized magnetic field satisfies $\nabla\times\mathbf E=jk\mathbf M$, $\nabla\times\mathbf M=-jk\epsilon_r\mathbf E$. Taking the tangential components of Maxwell's equations, applying the jump operator across the mean interface, and using the zeroth-order boundary conditions together with $[M_z^0]=0$, we obtain
\begin{align}
    [\partial_z\mathbf E_\parallel^{(0)}]
    &=\nabla_\parallel[E_z^{(0)}],\\
    [\partial_z\mathbf M_\parallel^{(0)}]
    &=-jk(\epsilon-1)\mathbf z\times\mathbf E_\parallel^{(0)}.
\end{align}
Substituting these relations into the first-order boundary conditions (\ref{eq:e2}) and (\ref{eq:m2}) yields
\begin{align}
    [\mathbf E_\parallel^{(1)}]&=\-\nabla_\parallel(h[E_z^{(0)}]),\label{eq:A4.1}\\
    [\mathbf M_\parallel^{(1)}]&=jk(\epsilon-1)h\mathbf z\times \mathbf E_\parallel^{(0)}.\label{eq:A4.2}
\end{align}

\subsection{Angular-Spectrum Amplitudes}
To determine the outgoing angular-spectrum amplitudes of the first-order scattered field, we transform the first-order boundary conditions into the transverse Fourier domain. Applying the transverse Fourier transform to (\ref{eq:A4.1}) and (\ref{eq:A4.2}) yields the boundary-field jumps in the Fourier domain, where $\tilde h_A(\Delta\mathbf K)$ denotes the Fourier transform of the surface-height profile over $A$. For an incident polarization $\alpha$, we define
\begin{align}
W&=jk(\epsilon-1)E_0\tilde h_A(\Delta\mathbf K), \quad X=\mathbf t_s\cdot\mathbf e_{\parallel,\alpha}, \notag\\
Y&=\mathbf s_s\cdot\mathbf e_{\parallel,\alpha},\quad Z=e_{z,\alpha}^{-}.
\end{align}
The resulting jumps in the Fourier-domain boundary fields are
\begin{align}
[\widetilde{\mathbf E}_{\parallel}^{(1)}]&=-\sin\theta_s WZ\mathbf t_s,\label{50}\\
[\widetilde{\mathbf M}_{\parallel}^{(1)}]&=W(-Y\mathbf t_s+X\mathbf s_s).\label{51}
\end{align}

Following the angular-spectrum decomposition in (\ref{E_total}), the tangential field components in the air half-space follow from $\mathbf p_s=-\cos\theta_s\mathbf t_s+\sin\theta_s\mathbf z$ and the normalized magnetic-field relation $\mathbf M=\mathbf k_s^+\times\mathbf E$, where $\mathbf k_s^+=\mathbf k_s^+/k$. The resulting tangential fields in the two half-spaces are
\begin{align}
\widetilde{\mathbf E}_{\parallel,\alpha}^{(1),+}&=a_{\alpha s}\mathbf s_s-\cos\theta_sa_{\alpha p}\mathbf t_s,\label{52}\\
\widetilde{\mathbf E}_{\parallel,\alpha}^{(1),-}&=a_{\alpha s}\mathbf s_s+(\sin\theta_s WZ-\cos\theta_s a_{\alpha p})\mathbf t_s,\label{53}\\
\widetilde{\mathbf M}_{\parallel,\alpha}^{(1),+}&=-\cos\theta_s a_{\alpha s}\mathbf t_s-a_{\alpha p}\mathbf s_s,\label{54}\\
\widetilde{\mathbf M}_{\parallel,\alpha}^{(1),-}&=\sqrt{\epsilon-\sin^2\theta_s}a_{\alpha s}\mathbf t_s\notag\\
&-\frac{\epsilon(W\sin\theta_sZ-\cos\theta_sa_{\alpha p})}{\sqrt{\epsilon-\sin^2\theta_s}}\mathbf s_s.\label{55}
\end{align}
Solving the system in (\ref{50})--(\ref{55}) by matching the corresponding components along $\mathbf s_s$ and $\mathbf t_s$ yields the outgoing angular-spectrum amplitudes:
\begin{align}
a_{\alpha s}&=\frac{WY}{\cos\theta_s+\sqrt{\epsilon-\sin^2\theta_s}},\label{56}\\
a_{\alpha p}&=\frac{W(\epsilon \sin\theta_s Z-X\sqrt{\epsilon-\sin^2\theta_s})}{\epsilon\cos\theta_s+\sqrt{\epsilon-\sin^2\theta_s}}.\label{57}
\end{align}

Substituting the zeroth-order field components from (\ref{33})--(\ref{35}) into the definitions of $X$, $Y$, and $Z$ gives their expressions for each incident polarization. Substituting these expressions into (\ref{56}) and (\ref{57}) and factoring out the common terms yields a unified expression for the four polarization-channel scattering amplitudes:
\begin{equation}
a_{\alpha\beta}(\mathbf K_s)=2jk\cos\theta_iE_0\tilde h_A(\Delta\mathbf K)q_{\alpha\beta}.\label{58}
\end{equation}
The corresponding polarization-dependent scattering factors are given in (\ref{59})--(\ref{62}). Finally, substituting (\ref{58}) into the far-field expression in (\ref{26}) gives the far-field electric-field component for incident polarization $\alpha$ and outgoing polarization $\beta$:
\begin{equation}
E_{\alpha\beta}^{(1)}(r)=\frac{k^2\cos\theta_i\cos\theta_s}{\pi r}E_0\tilde h_A(\Delta\mathbf K)q_{\alpha\beta}e^{jkr}.\label{63}
\end{equation}

\begin{figure*}[!t]
\centering
\begin{align}
q_{ss}&=\frac{(\epsilon-1)\cos\Delta\phi}{\left(\cos\theta_i+\sqrt{\epsilon-\sin^2\theta_i}\right)\left(\cos\theta_s+\sqrt{\epsilon-\sin^2\theta_s}\right)},\label{59}\\
q_{sp}&=-\frac{(\epsilon-1)\sin\Delta\phi\sqrt{\epsilon-\sin^2\theta_s}}{\left(\cos\theta_i+\sqrt{\epsilon-\sin^2\theta_i}\right)\left(\epsilon\cos\theta_s+\sqrt{\epsilon-\sin^2\theta_s}\right)},\label{60}\\
q_{ps}&=-\frac{(\epsilon-1)\sin\Delta\phi\sqrt{\epsilon-\sin^2\theta_i}}{\left(\epsilon\cos\theta_i+\sqrt{\epsilon-\sin^2\theta_i}\right)\left(\cos\theta_s+\sqrt{\epsilon-\sin^2\theta_s}\right)},\label{61}\\
q_{pp}&=\frac{(\epsilon-1)\left(\epsilon\sin\theta_i\sin\theta_s-\sqrt{\epsilon-\sin^2\theta_i}\sqrt{\epsilon-\sin^2\theta_s}\cos\Delta\phi\right)}{\left(\epsilon\cos\theta_i+\sqrt{\epsilon-\sin^2\theta_i}\right)\left(\epsilon\cos\theta_s+\sqrt{\epsilon-\sin^2\theta_s}\right)}.\label{62}
\end{align}
\vspace{-0.5em}
\noindent\rule{\textwidth}{0.4pt}
\end{figure*}

\subsection{Effective BRDF Formulation}
Having derived the far-field expression of the first-order scattered electric field, we now derive the polarization-resolved BRDF. Let $A$ denote the illuminated area on the mean surface plane. For an incident plane wave with complex amplitude $E_0$, the total incident power intercepted by the surface is
\begin{equation}
P_i=\frac{|E_0|^2A\cos\theta_i}{2Z_0} .
\end{equation}

For an incident polarization $\alpha$ and an outgoing polarization $\beta$, the differential scattered power per unit solid angle is obtained from the far-field Poynting flux as
\begin{equation}
\frac{dP_{\alpha\beta}}{d\Omega_s}=\frac{r^2}{2Z_0}\left\langle\left|E_{\alpha\beta}^{(1)}(r)\right|^2\right\rangle_h,
\end{equation}
where $\langle\cdot\rangle_h$ denotes the ensemble average over surface-height realizations. The BRDF is defined as the ratio of the outgoing radiance to the incident irradiance. Substituting the far-field expression in (\ref{59}) into the above power relation and evaluating the ensemble average using the two-dimensional surface-height PSD yields the polarization-resolved BRDF:
\begin{align}
\mathrm{BRDF}_{\alpha\beta}&=\frac{1}{P_i\cos\theta_s}\frac{dP_{\alpha\beta}}{d\Omega_s}\notag\\
&=\frac{16\pi^2}{\lambda^4}\cos\theta_i\cos\theta_s|q_{\alpha\beta}|^2S_h(f_x,f_y).\label{66}
\end{align}

To obtain a closed-form BRDF expression, we adopt an analytical model for the surface PSD. Specifically, we use an isotropic Gaussian PSD, a commonly used idealized model in rough-surface scattering studies. This model characterizes the surface roughness using two parameters and gives the following analytical expression:
\begin{align}
&S_h(f_x,f_y)=\pi\sigma^2l^2e^{-\left(\frac{\pi l f}{\lambda}\right)^2},\\
f&=\sqrt{\sin^2\theta_s+\sin^2\theta_i-2\sin\theta_s\sin\theta_i\cos\Delta\phi},
\end{align}
where $\sigma$ denotes the root-mean-square surface height, and $l$ denotes the lateral correlation length.

The BRDF in (\ref{66}) is polarization-resolved and depends on the incident and outgoing polarization pair. In practical multi-bounce scattering scenarios, the polarization state of the electromagnetic wave may change after each reflection due to polarization conversion. Explicitly tracking these changes across successive interactions would increase the computational complexity. We therefore introduce a polarization-averaged effective BRDF to provide a unified characterization of the scattering response without explicitly tracking the polarization state at each interaction:
\begin{equation}
|\bar{Q}|^2 = \frac{1}{2}\left(|q_{ss}|^2+|q_{sp}|^2+|q_{ps}|^2+|q_{pp}|^2 \right).
\end{equation}

Direct evaluation of this expression using (\ref{59})--(\ref{62}), however, requires computing four polarization-dependent scattering coefficients. Repeating these computations at each interaction in multi-bounce scattering scenarios can be cumbersome. To reduce this complexity, we adopt the geometric--Fresnel approximation proposed by Butler~\cite{Butler:15}. By comparing the microfacet BRDF with the modified Beckmann--Kirchhoff (MBK) BRDF, Butler derived an approximate relationship between the geometric factor and the four-channel polarization factor:
\begin{align}
    Q&=\frac{4u^4}{\cos\theta_i \cos\theta_s(\cos\theta_i+\cos\theta_s)^2}F(\theta_d)\\
    u^2&=\frac12(1+\cos\theta_i\cos\theta_s-\sin\theta_i \sin\theta_s \cos\Delta\phi)
\end{align}
where $u=\cos\theta_d$. This approximation replaces the four-channel computation with a single scalar expression. However, evaluating the exact Fresnel reflectance still requires the refractive index or relative permittivity of the material. To avoid explicitly specifying these quantities, we approximate the Fresnel reflectance using the Schlick formula:
\begin{align}
F(\theta_d)&=F(u,a,m)=F_0+(1-F_0)(1-u)^5,\\
F_0&=(1-m)0.04+ma
\end{align}
where $m$ denotes the metallic parameter, and $a$ denotes the diffuse albedo. To facilitate the integration of the proposed BRDF into 3D geometric representations, we express the effective BRDF in world coordinates using the surface normal $\mathbf n$ and the incident and outgoing directions $\boldsymbol\omega_i$ and $\boldsymbol\omega_s$. Following the standard BRDF convention, both directions are defined as unit vectors pointing away from the surface. Accordingly, the polarization-averaged effective BRDF in world coordinates is given by
\begin{equation}
f_r(\mathbf n,\boldsymbol\omega_i,\boldsymbol\omega_s)=\frac{64\pi^3\sigma^2l^2u^4F(u,a,m)e^{-\left(\frac{\pi lf}{\lambda}\right)^2}}{\lambda^4[\mathbf n\cdot(\boldsymbol\omega_i+\boldsymbol\omega_s)]^2},
\end{equation}
where
\begin{align}
u=&\sqrt{\frac{1+\boldsymbol\omega_i\cdot\boldsymbol\omega_s}{2}},\\
f=&\|(\mathbf I-\mathbf n\mathbf n^\top)(\boldsymbol\omega_i+\boldsymbol\omega_s)\|,\\
F(u,a,m)=&F_0+(1-F_0)(1-u)^5.
\end{align}
This formulation eliminates the dependence on local angular coordinates and enables direct BRDF evaluation using the surface normal and incident and outgoing directions in world coordinates.

\section{BRDF Validity and Robustness}\label{app:range}
The proposed BRDF is not intended to provide an exact electromagnetic solution to surface scattering. Instead, it is a physics-informed effective model guided by electromagnetic scattering theory. The first-order SPM provides its theoretical foundation, while learnable parameters allow adaptation to complex environments and compensate for the limitations of the approximation. To maintain physical consistency, we constrain the dimensionless roughness parameters $k\sigma$ and $kl$ to the nominal validity range of the first-order SPM in our main experiments. This constraint serves as a physically motivated prior for BRDF learning.

To assess the effect of this constraint, we remove the explicit bounds on $\sigma$ and $l$ and retrain the model under otherwise identical settings. The resulting performance changes only marginally, indicating that the proposed BRDF is not strongly dependent on these constraints. We further examine the distributions of the learned roughness parameters across all Gaussian primitives to assess their consistency with the nominal SPM validity range.

\begin{figure}[t]
    \centering
    \includegraphics[width=\columnwidth]{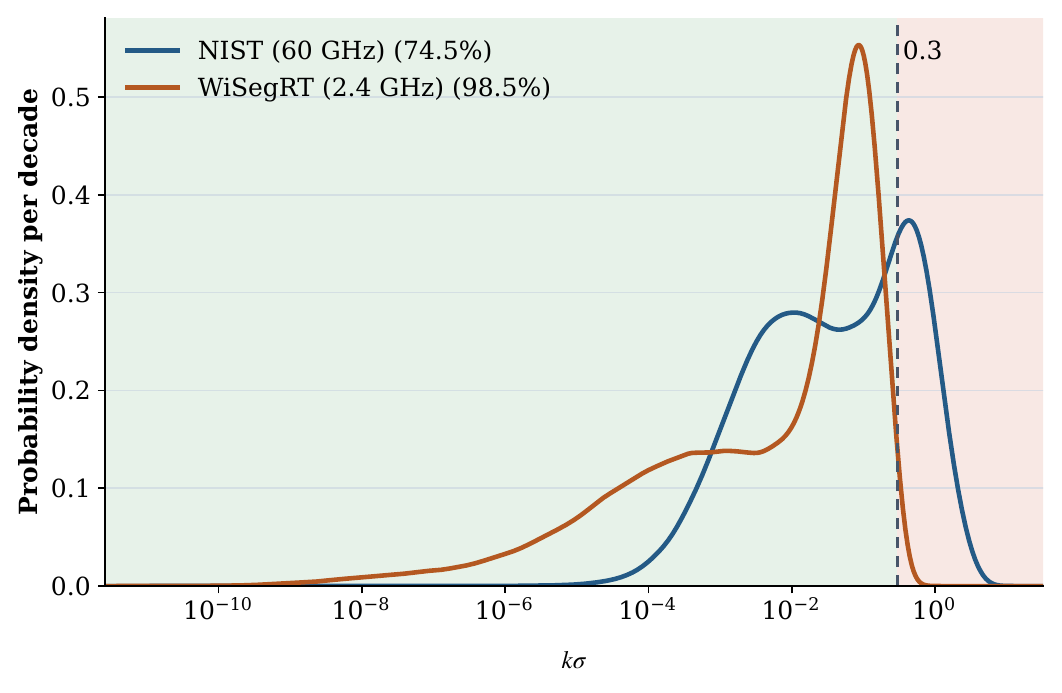}

    \vspace{2mm}

    \includegraphics[width=\columnwidth]{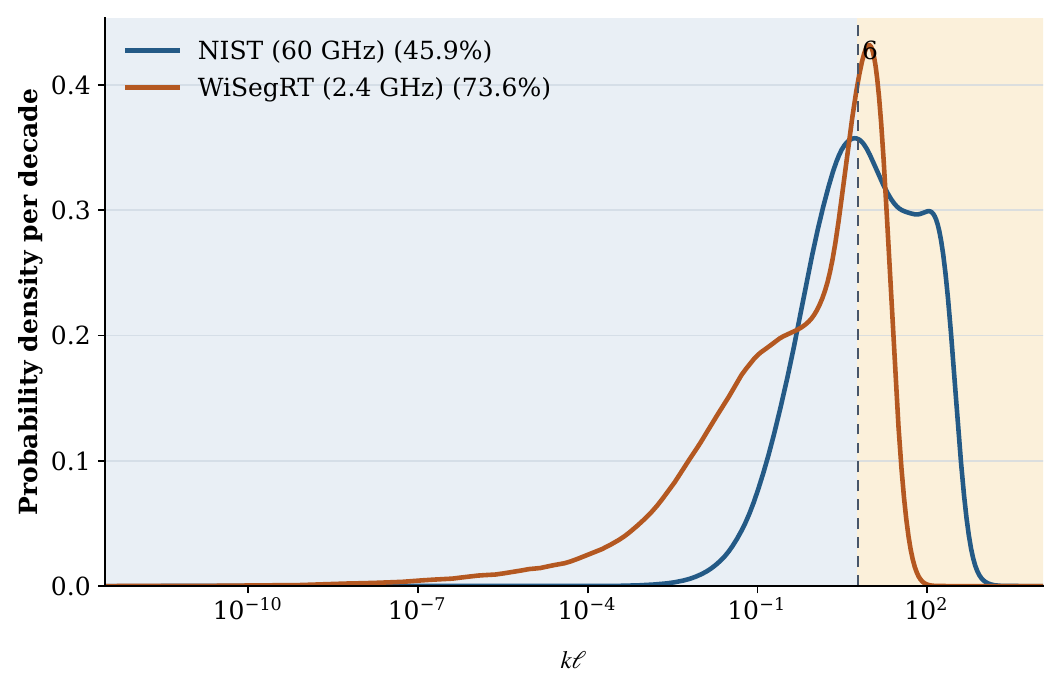}

    \caption{Distributions of the dimensionless RMS height $k\sigma$ and correlation length $kl$ across Gaussian primitives for the WiSegRT dataset and the NIST dataset. The dashed lines indicate the nominal validity thresholds of the first-order SPM.}
    \label{fig:roughness}
\end{figure}

Fig.~\ref{fig:roughness} presents the probability density functions of the dimensionless RMS height $k\sigma$ and correlation length $kl$ across all Gaussian primitives. At 2.4 GHz, 98.5\% of the Gaussian primitives fall within the nominal validity threshold for $k\sigma$, while 73.6\% fall within that for $kl$. These results indicate that most learned roughness parameters lie within the respective SPM validity limits, supporting the physical motivation of the proposed BRDF parameterization.

In contrast, at 60 GHz, a larger proportion of the learned parameters falls outside the nominal SPM validity range, particularly for $kl$. Nevertheless, removing the explicit parameter constraints results in only minor performance changes. This demonstrates the robustness of the proposed BRDF parameterization across the two frequency bands, even when some learned parameters extend beyond the nominal validity range of the first-order SPM.

\section{Ablation Study on Number of Rays}\label{app:sample}
\begin{table}[t]
    \centering
    \caption{Ablation study on the number of sampled rays.}
    \label{tab:sample}
    \small
    \renewcommand{\arraystretch}{0.95}
    \begin{tabularx}{\columnwidth}{c*{8}{>{\centering\arraybackslash}X}}
        \toprule
        & \multicolumn{4}{c}{NIST (60 GHz)}
        & \multicolumn{4}{c}{WisegRT (2.4 GHz)} \\
        \cmidrule(lr){2-5}
        \cmidrule(lr){6-9}
        $N_s$
        & MAE$\downarrow$ & PSNR$\uparrow$ & SSIM$\uparrow$ & Time$\downarrow$
        & MAE$\downarrow$ & PSNR$\uparrow$ & SSIM$\uparrow$ & Time$\downarrow$ \\
        \midrule
        4
        & 5.528 & 19.92 & 0.703 & 0.29
        & 15.304 & 16.45 & 0.702 & 0.60 \\

        8
        & 5.423 & 20.04 & 0.712 & 0.43
        & 15.257 & 16.49 & 0.705 & 0.80 \\

        12
        & 5.370 & 20.06 & 0.717 & 0.58
        & 15.233 & 16.51 & 0.706 & 0.91 \\

        20
        & 5.321 & 20.12 & 0.721 & 0.88
        & 15.216 & 16.55 & 0.707 & 1.18 \\

        32
        & 5.282 & 20.15 & 0.725 & 1.44
        & 15.208 & 16.58 & 0.707 & 1.69 \\
        \bottomrule
    \end{tabularx}
\end{table}

To evaluate the hemispherical integral in (\ref{eq:two_bounce}), we employ Monte Carlo sampling to obtain a numerical approximation. Specifically, by drawing $N_s$ independent samples from a probability density function $p(\boldsymbol{\omega})$, the integral can be approximated as
\begin{equation}
I \approx \frac{1}{N_s}\sum_{i=1}^{N_s}\frac{f(\boldsymbol{\omega}_i)}{p(\boldsymbol{\omega}_i)},\quad \boldsymbol{\omega}_i \sim p(\boldsymbol{\omega}).
\end{equation}
where $N_s$ denotes the number of Monte Carlo samples, and $p(\boldsymbol{\omega})$ represents the corresponding sampling probability density function.

As shown in Tab.~\ref{tab:sample}, increasing the number of Monte Carlo samples generally leads to improved reconstruction quality. However, the marginal improvements gradually diminish as the sampling budget increases. This observation highlights the trade-off between reconstruction quality and computational efficiency: although increasing the number of samples yields more accurate estimates of the hemispherical integral, the resulting quality gains become progressively smaller relative to the additional computational cost. Therefore, $N_s=8$ provides a reasonable balance between reconstruction quality and computational efficiency, whereas $N_s=32$ achieves slightly better reconstruction quality at a substantially higher computational cost.

\section{Ablation Study on Gaussian convolution}\label{app:gaussian}
\begin{table}[t]
    \centering
    \caption{Ablation study on Gaussian convolution.}
    \label{tab:conv}
    \small
    \renewcommand{\arraystretch}{0.9}
    \begin{tabularx}{\columnwidth}{l*{6}{>{\centering\arraybackslash}X}}
    \toprule
    & \multicolumn{3}{c}{NRS}
    & \multicolumn{3}{c}{Relighting (Zero-Shot)} \\
    \cmidrule(lr){2-4}
    \cmidrule(lr){5-7}
    & MAE$\downarrow$ & PSNR$\uparrow$ & SSIM$\uparrow$
    & MAE$\downarrow$ & PSNR$\uparrow$ & SSIM$\uparrow$ \\
    \midrule
    w/o conv.
    & 5.940 & 19.45 & 0.618
    & 10.783 & 16.54 & 0.624 \\

    Ours
    & \textbf{5.423} & \textbf{20.04} & \textbf{0.712}
    & \textbf{8.040} & \textbf{17.54} & \textbf{0.694} \\

    \bottomrule
    \end{tabularx}
\end{table}

In practical communication systems, the angular resolution of a receiving antenna is inherently limited. Consequently, signals arriving from different directions within the main lobe may contribute to the same angular response. To model this angular aggregation effect, we apply a fixed $3\times 3$ Gaussian convolution to the accumulated linear-amplitude map prior to conversion to the dB domain. 

Tab.~\ref{tab:conv} reports the quantitative comparison between the baseline without Gaussian convolution and the proposed method on NRS and zero-shot relighting. The results demonstrate that Ours consistently outperforms the baseline across all evaluation metrics in both settings. In particular, the improvements are more pronounced in zero-shot relighting, where the MAE decreases from 10.783 to 8.040, corresponding to a relative reduction of 25.4\%.  These results demonstrate the effectiveness of modeling the finite angular resolution of the receiving antenna, leading to consistent improvements in reconstruction quality across both evaluation settings and particularly substantial gains in MAE under zero-shot relighting.

\section{Ablation Study on Supervision Domain}\label{app:db}
Although linear-domain supervision directly preserves the original signal magnitudes, it can be suboptimal when handling signals with a wide dynamic range. Specifically, since pointwise errors in the linear domain are measured in absolute terms rather than relative terms, errors associated with high-amplitude regions can dominate the overall optimization objective, while relative deviations in low-amplitude regions receive comparatively little emphasis. Consequently, the network may prioritize the reconstruction of high-intensity components at the expense of fine details in low-intensity regions. To address this issue, we adopt dB-domain supervision, which maps the original signal into a logarithmic space and compresses its dynamic range, thereby providing a more balanced representation across different signal magnitudes.

A direct conversion from linear amplitude to the dB domain is defined as $D(x)=10\log_{10}(x)$, with a derivative of $D'(x)=10/(x\ln 10)$. As $x$ approaches zero, the gradient magnitude increases rapidly, potentially amplifying numerical noise and leading to unstable optimization. To mitigate this issue, we employ a shifted logarithmic transformation:
\begin{equation}
    D(x)=10\log_{10}\left(1+\frac{x}{x_0}\right),
\end{equation}
where $x_0>0$ denotes a reference scale. Based on this transformation, the supervision objective is formulated as
\begin{align}
    \mathcal{L}_e&=\lambda_e \mathcal{L}_1\left(D(\hat{x}),D(x)\right)\notag\\
    &+(1-\lambda_e)\mathcal{L}_{\text{SSIM}}\left(D(\hat{x}),D(x)\right).
\end{align}
Here, $x$ and $\hat{x}$ denote the ground-truth and predicted amplitudes, respectively.

\begin{figure*}[t]
    \centering
    \includegraphics[width=\textwidth]{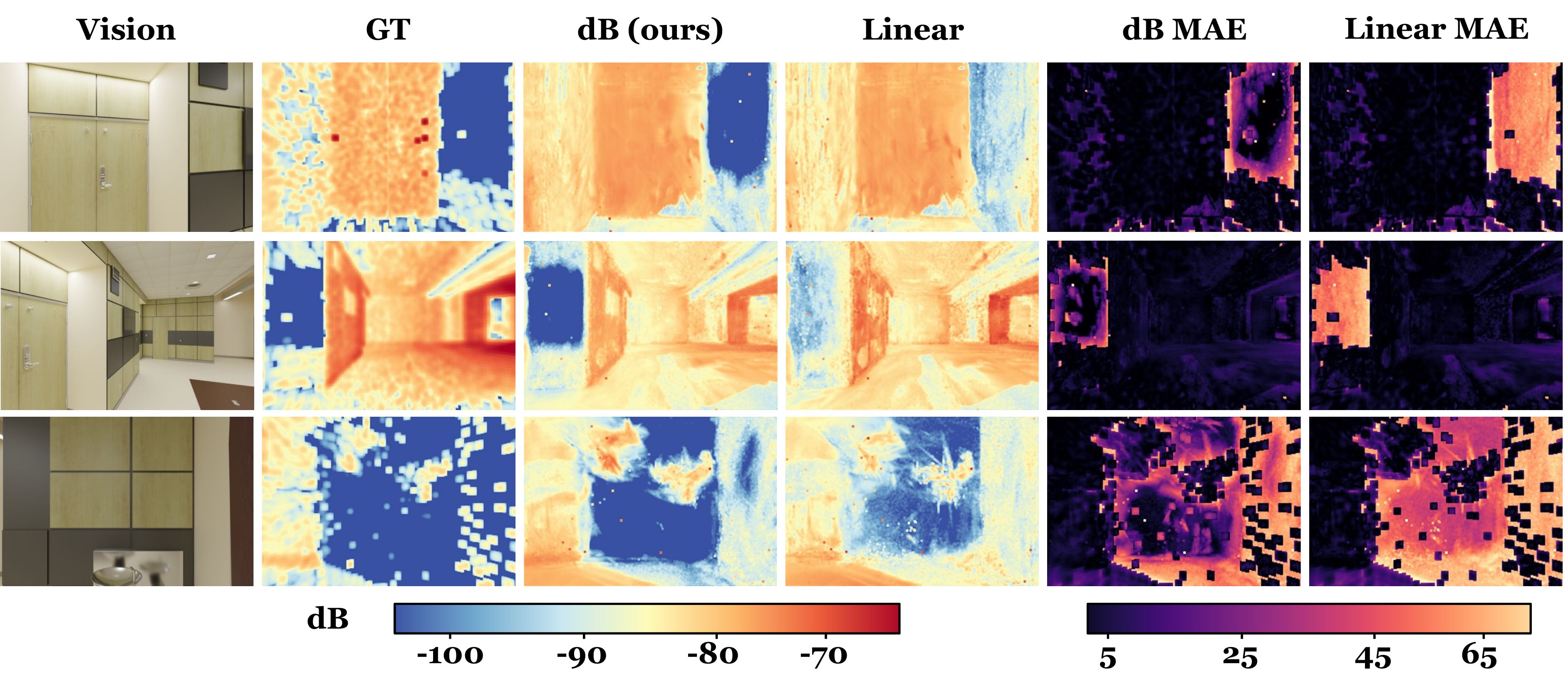}
    \caption{Visual comparisons of APS and MAE maps under dB-domain and linear-domain supervision. For dB-domain supervision, the GT values are converted to the dB domain before training, whereas linear-domain supervision uses the original GT values during training. In both cases, the predictions are converted to the dB domain for visualization.}
    \label{fig:db}
\end{figure*}

Tab.~\ref{tab:db} presents a quantitative comparison between linear-domain and dB-domain supervision. The results show that dB-domain supervision consistently outperforms linear-domain supervision across all six evaluation metrics. The qualitative comparisons in Fig.~\ref{fig:db} further illustrate the advantages of the proposed supervision strategy. Although both configurations recover the overall spatial structure of the ground-truth signal, the Linear MAE maps exhibit more pronounced localized errors in several low-amplitude regions. In contrast, the dB MAE maps show weaker error responses in these regions. This observation suggests that linear-domain supervision provides insufficient emphasis on weak signal components, whereas logarithmic transformation helps alleviate this limitation by compressing the dynamic range and balancing the representation of signals with different amplitudes. Taken together, the quantitative and qualitative results demonstrate the effectiveness of dB-domain supervision in improving reconstruction accuracy, particularly in low-amplitude regions.

\begin{table}[t]
    \centering
    \caption{Ablation study on the supervision domain. The predictions are converted to the dB domain.}
    \label{tab:db}
    \small
    \renewcommand{\arraystretch}{0.9}
    \begin{tabularx}{\columnwidth}{l*{6}{>{\centering\arraybackslash}X}}
        \toprule
        & \multicolumn{3}{c}{NRS}
        & \multicolumn{3}{c}{Relighting (Zero-shot)} \\
        \cmidrule(lr){2-4}
        \cmidrule(lr){5-7}
        & MAE$\downarrow$ & PSNR$\uparrow$ & SSIM$\uparrow$
        & MAE$\downarrow$ & PSNR$\uparrow$ & SSIM$\uparrow$ \\
        \midrule
        Linear domain
        & 5.75 & 19.37 & 0.668
        & 9.83 & 16.77 & 0.683 \\

        dB domain
        & \textbf{5.423} & \textbf{20.04} & \textbf{0.712}
        & \textbf{8.04} & \textbf{17.54} & \textbf{0.694} \\
        \bottomrule
    \end{tabularx}
\end{table}

\begin{figure*}[t]
    \centering
    \includegraphics[width=0.8\textwidth]{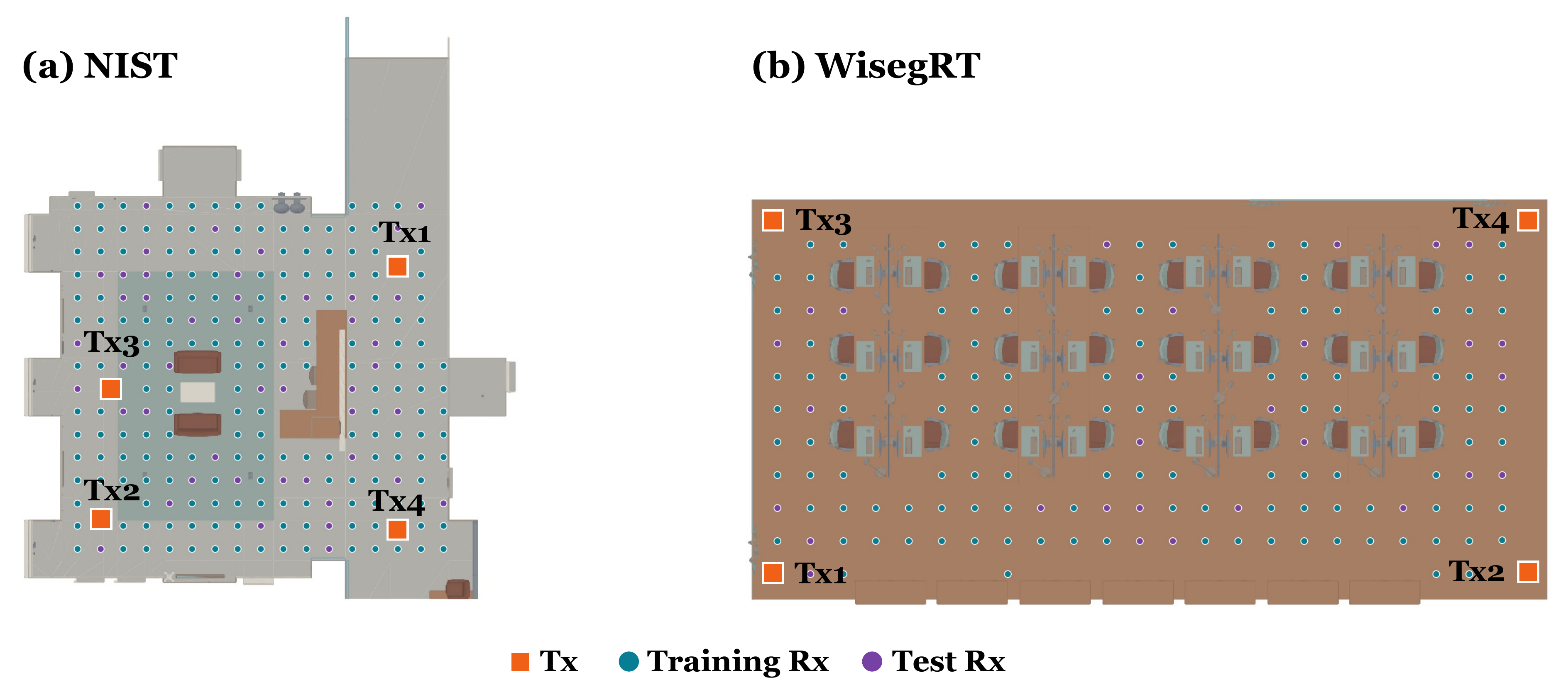}
    \caption{Visual comparisons of WRGS with SOTA methods on NRS and RF relighting. For each scene, the leftmost column (Setting) shows the receiver-view image and the corresponding floor plan, with the locations of Tx and the Rx indicated. NRS is evaluated using Tx1, whereas RF relighting is evaluated using Tx2.}
    \label{fig:map}
\end{figure*}

\section{Dataset Detail}\label{app:data}
We consider two indoor environments for RF simulation: the NIST lobby model released with RF-3DGS and Scene06 from WiSegRT. Fig.~\ref{fig:map} illustrates the spatial layouts of both environments, including the Tx locations and Rx sampling positions. Four transmitters are deployed in each environment, with receiver locations sampled throughout the corresponding indoor areas.

\noindent\textbf{NIST.} The NIST lobby represents a relatively complex indoor environment with open areas, connected corridors, and indoor furnishings. RF observations are generated at a carrier frequency of 60~GHz, with metal and gypsum board as the two predominant materials.

\noindent\textbf{WiSegRT.} WiSegRT provides indoor 3D environments with detailed object-level and material-level representations. We select Scene06, which covers a spatial region of $16\times 8\times 2.4$~m, corresponding to a horizontal floor area of approximately $128$~m$^2$. RF observations are generated at 2.4~GHz. The scene contains regularly arranged indoor facilities, with plastic and metal as the dominant materials.

Following the pipeline of RF-3DGS~\cite{11355734}, we use NVIDIA Sionna RT to simulate RF propagation in the 3D scenes. Given the scene geometry, material properties, and Tx/Rx configurations, ray tracing generates multipath components (MPCs) with corresponding amplitudes, angles of arrival (AoAs), angles of departure (AoDs), and propagation delays. We construct a two-dimensional angular-domain RF map by accumulating the linear-amplitude contributions of MPCs according to their AoAs.

\begin{figure}[t]
    \centering
    \includegraphics[width=\columnwidth]{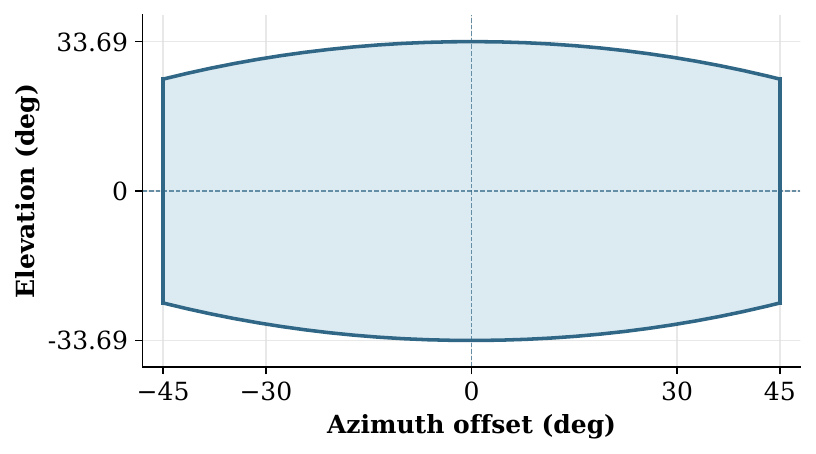}
    \caption{Actual angular coverage of the pinhole camera.}
    \label{fig:view}
\end{figure}

We then project the angular-domain RF maps onto a pinhole camera plane to obtain image-like RF observations, where each pixel represents a viewing ray and records the RF response along the corresponding direction. For each Rx position, four views are generated at yaw angles of $-90^\circ$, $0^\circ$, $90^\circ$, and $180^\circ$, with a fixed pitch angle of $0^\circ$. Each view has a $90^\circ$ field of view (FoV), and the corresponding angular coverage is illustrated in Fig.~\ref{fig:view}. The dataset configurations and splits are summarized in Tab.~\ref{tab:settings}.

The electromagnetic properties of the materials in both environments are summarized in Tab.~\ref{tab:material}. The relative permittivity is kept consistent across shared material categories, whereas the electrical conductivity is specified according to the corresponding simulation frequency.

\begin{table}[t]
    \centering
    \caption{Simulation configurations shared across both scenarios and scenario-specific settings.}
    \label{tab:settings}
    \small
    \renewcommand{\arraystretch}{0.9}
    \begin{tabularx}{\columnwidth}{>{\raggedright\arraybackslash}X*{2}{>{\centering\arraybackslash}X}}
        \toprule
        \multirow[c]{2}{*}{Parameter}
        & \multicolumn{2}{c}{Value} \\
        \cmidrule(lr){2-3}
        & NIST & WiSegRT\\
        \midrule
        Tx array
        & \multicolumn{2}{c}{1$\times$1, 1 element} \\
        Rx array
        & \multicolumn{2}{c}{1$\times$1, 1 element} \\
        Element spacing
        & \multicolumn{2}{c}{$0.5\lambda$} \\
        Radiation pattern
        & \multicolumn{2}{c}{Isotropic} \\
        Polarization
        & \multicolumn{2}{c}{vertical} \\
        Tx power
        & \multicolumn{2}{c}{44 dBm} \\
        Interaction depth
        & \multicolumn{2}{c}{8} \\
        Line-of-sight
        & \multicolumn{2}{c}{True} \\
        Reflection
        & \multicolumn{2}{c}{True} \\
        Scattering
        & \multicolumn{2}{c}{True} \\
        Diffraction
        & \multicolumn{2}{c}{True} \\
        Training set
        & \multicolumn{2}{c}{800 Rx $\times$ 4 views} \\
        Validation set
        & \multicolumn{2}{c}{125 Rx $\times$ 4 views} \\
        View directions
        & \multicolumn{2}{c}{$-90^\circ$, $0^\circ$, $90^\circ$, $180^\circ$} \\
        View resolution
        & \multicolumn{2}{c}{200$\times$300} \\

        \midrule

        Carrier frequency
        & 60 GHz & 2.4 GHz \\
        Tx1 coordinates
        & $(6.9, 0, 0.3)$
        & $(-7.5, -3.5, 2.2)$ \\
        Tx2 coordinates
        & $(-3.3, -9.2, 0.3)$
        & $(7.5, 3.5, 2.2)$ \\
        Tx3 coordinates
        & $(-1.0, -5.0, 0.3)$
        & $(-7.5, 3.5, 2.2)$ \\
        Tx4 coordinates
        & $(-3.0, -4.0, 0.3)$
        & $(7.5, -3.5, 2.2)$ \\

        \bottomrule
    \end{tabularx}
\end{table}

\begin{table}[t]
    \centering
    \caption{Relative permittivity and electrical conductivity of different materials. Gypsum board and ceiling board appear only in the NIST environment.}
    \label{tab:material}
    \small
    \renewcommand{\arraystretch}{0.9}
    \begin{tabularx}{\columnwidth}{l*{3}{>{\centering\arraybackslash}X}}
        \toprule
        \multicolumn{1}{c}{\multirow[c]{2}{*}{Material}}
        & \multicolumn{1}{c}{\multirow[c]{2}{*}{\makecell[c]{Relative\\Permittivity}}}
        & \multicolumn{2}{c}{Conductivity (S/m)} \\
        \cmidrule(lr){3-4}
        & & 60 GHz
        & 2.4 GHz \\
        \midrule
        Plastic & 2.30 & 0 & 0 \\
        Leather & 1.80 & 0 & 0 \\
        Fabric & 1.80 & 0 & 0 \\
        Wood & 1.99 & 0.378373 & 0.012012 \\
        Concrete & 5.24 & 3.719328 & 0.118073 \\
        Metal & 1.00 & $1.0 \times 10^{7}$ & $1.0 \times 10^{7}$ \\
        Glass & 6.31 & 0.866879 & 0.011629 \\
        Gypsum & 2.73 & 0.398100 & -- \\
        Ceiling & 1.48 & 0.089723 & -- \\
        \bottomrule
    \end{tabularx}
\end{table}

\section{Transmittance Approximation}\label{app:Transmission}
The transmittance of radio waves propagating through a material is governed by several physical factors, including the carrier frequency, the electromagnetic properties of the medium, and the material thickness along the propagation path. Accurately estimating transmittance therefore requires not only material-specific electromagnetic parameters but also geometric information, particularly the thickness of the intervening objects. However, conventional 3DGS primarily relies on appearance-based optimization to reconstruct scene geometry and appearance from visual observations, without explicitly estimating material-specific electromagnetic properties or the physical thickness of objects. Recovering these quantities would consequently require additional measurements or modeling assumptions, substantially increasing the complexity of the framework.

To avoid these additional requirements, we use the opacity parameter associated with each Gaussian primitive as a proxy for radio-wave transmittance. Since Gaussian opacity is naturally optimized during the 3DGS reconstruction process, this choice enables us to represent transmittance using the existing Gaussian parameters, without explicitly estimating material properties or object thickness. Although this approximation does not capture the exact physical transmittance, it provides a practical and computationally efficient alternative that is readily compatible with the 3DGS representation.

\section{Additional Visualizations}\label{app:visual}
\begin{figure*}[t]
    \centering
    \includegraphics[width=\textwidth]{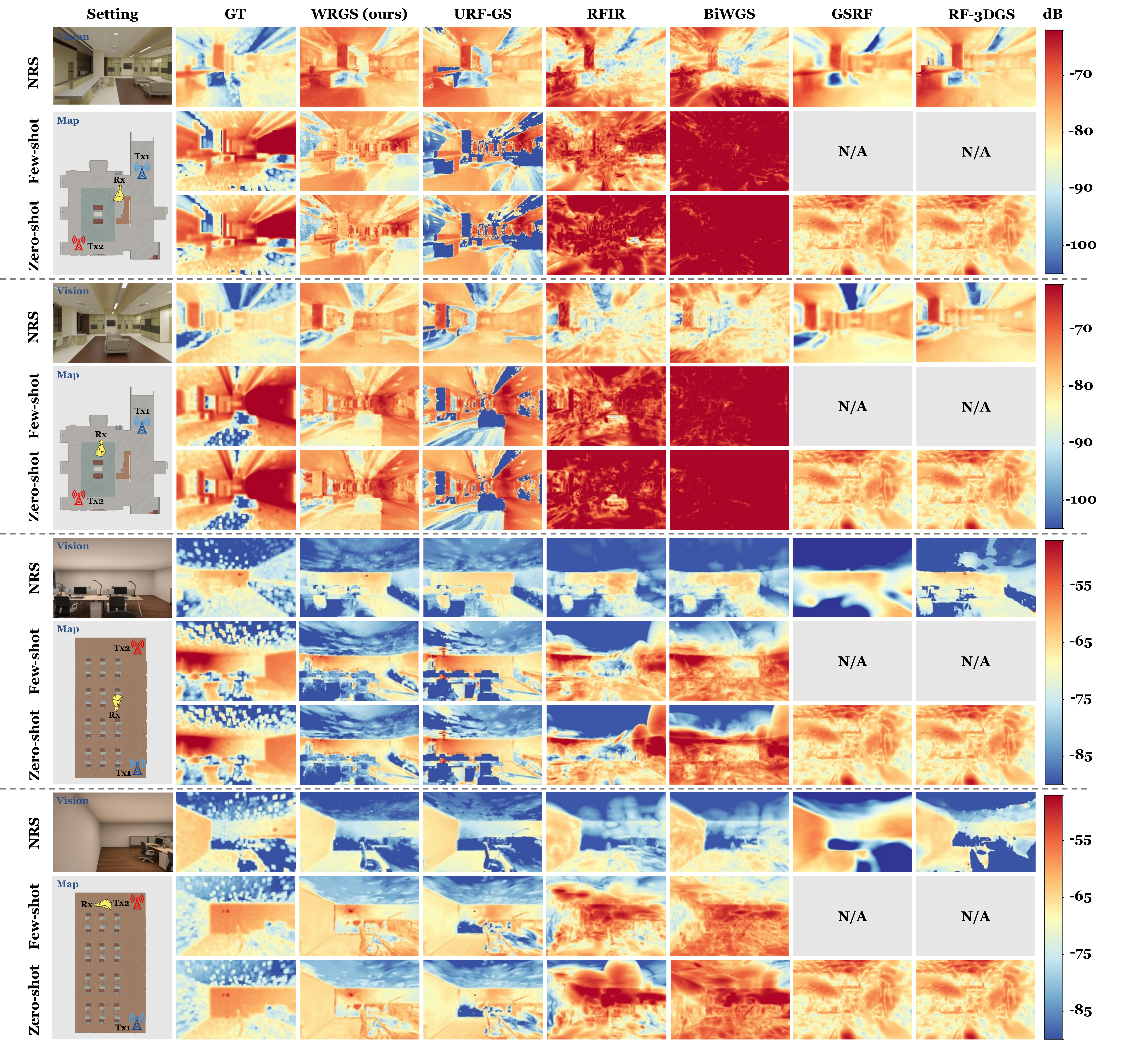}
    \caption{Visual comparisons of WRGS with SOTA methods on NRS and RF relighting. For each scene, the leftmost column (Setting) shows the receiver-view image and the corresponding floor plan, with the locations of Tx and the Rx indicated. NRS is evaluated using Tx1, whereas RF relighting is evaluated using Tx2. RF-3DGS and GSRF are evaluated only in the few-shot setting because they do not support zero-shot evaluation.}
    \label{fig:visual_s}
\end{figure*}

Fig.~\ref{fig:visual_s} presents additional qualitative comparisons of WRGS with SOTA methods across multiple scenes under both NRS and RF relighting settings. The results demonstrate that WRGS generates APS that more closely match the GT while better preserving the spatial distribution and local variations of received signal strength across different transmitter configurations.
